\documentclass[pdflatex,sn-nature]{sn-jnl}

\usepackage{graphicx}%
\usepackage{multirow}%
\usepackage{amsmath,amssymb,amsfonts}%
\usepackage{amsthm}%
\usepackage{mathrsfs}%
\usepackage[title]{appendix}%
\usepackage{xcolor}%
\usepackage[table]{xcolor}
\usepackage{textcomp}%
\usepackage{manyfoot}%
\usepackage{booktabs}%
\usepackage{algorithm}%
\usepackage{algorithmicx}%
\usepackage{algpseudocode}%
\usepackage{listings}%
\usepackage{siunitx}
\usepackage{hyperref}
\makeatletter
\let\orcidlogo\@undefined
\makeatother
\usepackage{orcidlink}
\usepackage{pgfplotstable}
\usepackage{rotating}
\pgfplotsset{compat=1.18}
\DeclareSIUnit{\foot}{ft}
\DeclareSIUnit{\cubicfoot}{ft^3}
\newcommand{\figurea}[1]{\hyperref[#1]{\ref*{#1}a}}
\newcommand{\figureb}[1]{\hyperref[#1]{\ref*{#1}b}}
\newcommand{\figurec}[1]{\hyperref[#1]{\ref*{#1}c}}
\newcommand{\figured}[1]{\hyperref[#1]{\ref*{#1}d}}
\newcommand{\figuree}[1]{\hyperref[#1]{\ref*{#1}e}}
\newcommand{\figuref}[1]{\hyperref[#1]{\ref*{#1}f}}
\newcommand{\figureg}[1]{\hyperref[#1]{\ref*{#1}g}}
\newcommand{\figureh}[1]{\hyperref[#1]{\ref*{#1}h}}
\newcommand{\figurei}[1]{\hyperref[#1]{\ref*{#1}i}}
\newcommand{\figurej}[1]{\hyperref[#1]{\ref*{#1}j}}

\theoremstyle{thmstyleone}%
\theoremstyle{thmstyletwo}%

\theoremstyle{thmstylethree}%

\begin{document}

\ifdefined\SUPPLEMENTARY

\setcounter{table}{0}
\setcounter{figure}{0}
\setcounter{equation}{0}
\renewcommand{\thetable}{\arabic{table}}
\renewcommand{\thefigure}{\arabic{figure}}
\renewcommand{\theequation}{S.\arabic{equation}}
\renewcommand{\figurename}{Supplementary Figure}
\renewcommand{\tablename}{Supplementary Table}
\newcommand{\suppsection}[1]{\section*{#1}\addcontentsline{toc}{section}{#1}}
\newcommand{\suppsubsection}[1]{\subsection*{#1}\addcontentsline{toc}{subsection}{#1}}
\clearpage
\pagenumbering{arabic}
\setcounter{page}{1}

\section*{Supplementary Information}
\vskip10pt
{\Titlefont \textbf{Physical-Variable-Informed Deep Learning for Robust Diagnosis in Industrial Fluid Transport Systems}\par}
\vskip15pt
{\Authorfont Yu Sha~\orcidlink{0000-0003-4521-2077}\textsuperscript{1,4}, Junqi Tao~\orcidlink{0000-0002-4406-2284}\textsuperscript{1,3}, Dixin Zhou~\orcidlink{0009-0006-5509-785X}\textsuperscript{2}, Horst Stoecker~\orcidlink{0000-0002-3282-3664}\textsuperscript{4,5,6}, Domagoj Vnucec~\orcidlink{0000-0002-2752-9612}\textsuperscript{7}, Andreas Widl~\orcidlink{0009-0009-0664-2024}\textsuperscript{7}, Kai Zhou~\orcidlink{0000-0001-9859-1758}\textsuperscript{1,2,4*}\par}
\vskip7pt
{\addressfont
\textsuperscript{1}School of Science and Engineering, The Chinese University of Hong Kong, Shenzhen, 518172, China.\par
\textsuperscript{2}School of Artificial Intelligence, The Chinese University of Hong Kong, Shenzhen, 518172, China.\par
\textsuperscript{3}Key Laboratory of Quark \& Lepton Physics and Institute of Particle Physics, Central China Normal University, Wuhan, 430079, China.\par
\textsuperscript{4}Frankfurt Institute for Advanced Studies, Frankfurt am Main, 60438, Germany.\par
\textsuperscript{5}Institut f{\"u}r Theoretische Physik, Goethe Universit{\"a}t Frankfurt, Frankfurt am Main, 60438, Germany.\par
\textsuperscript{6}GSI Helmholtzzentrum f{\"u}r Schwerionenforschung GmbH, Darmstadt, 64291, Germany.\par
\textsuperscript{7}SAMSON AG, Frankfurt am Main, 60314, Germany.\par
*Corresponding author(s). E-mail(s): zhoukai@cuhk.edu.cn;\par}
\label{sec: 10_Appendix}
\tableofcontents

\suppsection{Supplementary Methods}
\label{sec:supplementary_methods}
\suppsubsection{Definition of Cavitation Intensity}
\label{sec: 10_Definition of Cavitation Intensity}
Supplementary Figure \ref{fig: AppPressureDistribution} demonstrates the variation in local pressure in a one-dimensional flow. Cavitation does not occur and the valve functions normally when the minimum pressure ${p}_{min}$ exceeds the vapour pressure ${p}_{v}$. However, when ${p}_{v}>{p}_{min}$, cavitation begins. In practice, direct measurement is difficult because the minimum pressure is downstream of the restriction. Therefore, the cavitation coefficient ${X}_{fz}$ is proposed, which represents the ratio of the external pressure difference to the internal pressure difference. This coefficient can be determined empirically by assuming that cavitation noise begins when the minimum pressure ${p}_{min}$ matches the vapour pressure ${p}_{v}$. Consequently, the cavitation coefficient ${X}_{fz}$ can be inferred from noise measurements, whose inferred value varies with the valve load. The formulas for the cavitation coefficient ${X}_{fz}$ and the operating pressure ratio ${X}_{f}$ are given below:
\begin{equation}
{X}_{fz}=\frac{{p}_{u}-{p}_{d}}{{p}_{u}-{p}_{min}}, \,\,\,\,\,\,\,\,\,\,\,\,\,\,\,\,\,\,\,\,\,\,\,\,\,\,\,\, {X}_{f}=\frac{{p}_{u}-{p}_{d}}{{p}_{u}-{p}_{v}},
\end{equation}
where ${p}_{u}$ represents the upstream pressure, ${p}_{d}$ denotes the downstream pressure, ${p}_{min}$ refers to the minimum pressure within the valve and ${p}_{v}$ denotes the vapour pressure. 
\begin{figure*}[htbp]
    \centering
    \includegraphics[width=0.6\textwidth,height=40mm]{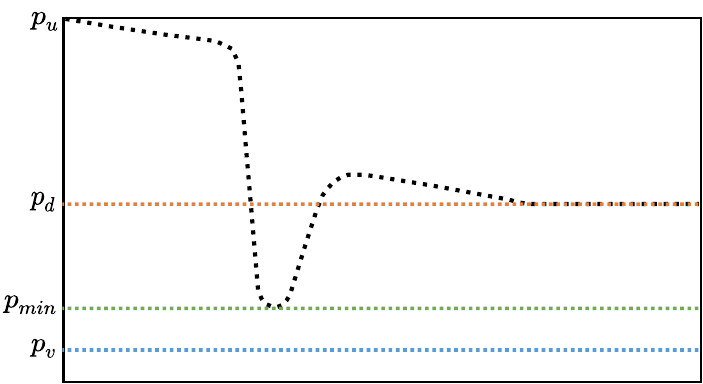}
    \caption{\textbf{Pressure flow model in a valve}. The solid black, dotted orange, dotted green and dotted blue lines represent the upstream pressure ${p}_{u}$, the downstream pressure ${p}_{d}$, the minimum pressure ${p}_{min}$ and the vapour pressure ${p}_{v}$, respectively.}
    \label{fig: AppPressureDistribution}
\end{figure*}

When all coefficients are determined across the entire range of valve openings, the following conclusions can be drawn:
\begin{itemize}
\item ${X}_{f}<{X}_{fz}$: The valve operates without cavitation and the flow may be either turbulent or laminar.
\item ${X}_{f}\geq {X}_{fz}$: When ${X}_{f}={X}_{fz}$, the valve exhibits incipient cavitation. As the difference between ${X}_{fz}$ and ${X}_{f}$ increases, the cavitation region expands due to the pressure drop caused by higher flow velocities.
\item ${X}_{f}>1$: In this case, the bubbles do not implode in the valve but rather continue to flow into the pipe because the downstream pressure ${p}_{d}$ is lower than the vapour pressure ${p}_{v}$. This phenomenon is called flashing.
\end{itemize}

The cavitation coefficient ${X}_{fz}$ is fluid-specific and is measured empirically. Its value varies for different liquid media due to changes in viscosity, dissolved gas content and other factors.

\suppsubsection{Cavitation Event Intensity Diagnosis}
\label{sec: Cavitation Event Intensity Diagnosis}
Cavitation is defined as the phenomenon involving the formation, growth and collapse of local bubbles or vapour cavities in a liquid. In piping systems, the acoustic signals of different flow conditions (different levels of cavitation or non-cavitation) are recorded as continuous waveforms using acoustic sensors. Each observed acoustic signal records the entire physical process from the beginning to the end of the corresponding flow state. In our experiments, cavitation intensity recognition mainly distinguishes incipient cavitation, constant cavitation, choked-flow cavitation and non-cavitation. Whether severe or subtle, any form of cavitation may indicate a potential issue or failure in system operation. Therefore, it is crucial for operators of industrial systems to effectively and precisely recognize different intensities of cavitation in order to implement appropriate countermeasures. 

\suppsubsection{Acoustic Signal Augmentation}
\label{sec: Acoustic Signals Augmentation}
Formally, each acoustic signal is represented as ${\mathbf{x}_{1:T}} = \{ {x_1},{x_2}, \ldots ,{x_T}\}  \in {{\mathbb{R}}^{T \times 1}}$. Given the intentionally maintained steady flow state (i.e., it is always the same flow-state class within each individual measurement lasting \SI{3}{s} or \SI{25}{s}) in each recorded stream and the high temporal resolution of the sensor, each signal can be split into several segments, each containing sufficient information for detection. The segment length is chosen to be sufficiently long to account for the inherent randomness in the noise emission, and each segment can be treated as an independent sample. Therefore, we apply a sliding-window technique $\mathcal{W}(\cdot)$ with window size $W_s$ and step size $S$ to divide the acoustic signal sequence into a set of subsequences ${\mathbf{X}_{1:W_s}} = \{ \mathbf{x}_{1:W_s}^1,\mathbf{x}_{1:W_s}^2, \ldots ,\mathbf{x}_{1:W_s}^M\}  \in {\mathbb{R}^{W_s \times M}}$, where $M = \left\lfloor {\frac{{T - W_s}}{S}} \right\rfloor  + 1$ is the number of subsequences. The sliding-window technique is a crucial part of the acoustic signal preprocessing.

\suppsubsection{Time-Frequency Transform}
\label{sec: Time-Frequency Transform}
The time-frequency (T-F) transform provides more detailed and comprehensive information across both time and frequency dimensions. In this work, the time-frequency representation is computed using the continuous wavelet transform (CWT), which is particularly suitable for non-stationary acoustic signal analysis because it captures localized signal characteristics at multiple scales. Formally, after applying the sliding-window operator $\mathcal{W}(\cdot)$, we obtain a set of subsequences ${\mathbf{X}_{1:W_s}} = \{ \mathbf{x}_{1:W_s}^1,\mathbf{x}_{1:W_s}^2, \ldots ,\mathbf{x}_{1:W_s}^M\} \in {\mathbb{R}^{W_s \times M}}$. For each subsequence $\mathbf{x}_{1:W_s}^i$, the CWT maps it into the time-frequency domain and is defined as:
\begin{equation}
\label{eq: CWT}
\mathbf{s}_i(a,b) = \frac{1}{\sqrt{|a|}} \sum\limits_{n=1}^{W_s} x_n^i \psi^*\left( \frac{n-b}{a} \right),
\end{equation}
where $\mathbf{s}_i(a,b)$ denotes the wavelet coefficient of the $i$-th subsequence at scale $a$ and translation $b$, $\psi(\cdot)$ is the mother wavelet, and $\psi^*(\cdot)$ represents its complex conjugate. The scale parameter $a$ controls the frequency resolution, while the translation parameter $b$ determines the temporal location of the wavelet. By varying $a$ and $b$, the CWT yields a multi-scale time-frequency representation for each acoustic subsequence. The essence of the CWT is to correlate the input signal with scaled and shifted versions of the mother wavelet, thereby revealing transient and non-stationary patterns across different frequency bands. In our method, the CWT plays an essential role in the preprocessing of the acoustic signal, where each subsequence $\mathbf{x}_{1:W_s}^i$ is transformed into a corresponding time-frequency spectrogram $\mathbf{s}_i$. These spectrograms are subsequently used as the input representations for the diagnosis framework.

\suppsubsection{Cavitation Datasets}
\label{sec: Cavitation Datasets}
The cavitation datasets are provided by SAMSON AG in Frankfurt. The acoustic signals are recorded under five distinct flow conditions, generated by adjusting the differential pressure at various constant upstream pressures of the control valve. These conditions include choked-flow cavitation, constant cavitation, incipient cavitation, turbulent flow and no-flow state (see Supplementary Table \ref{tab: CavitationDatasets-FlowStatus}). The detailed training and test statistics for the three real-world cavitation datasets with or without data augmentation are provided in Supplementary Table \ref{tab: training and test sets}. 
\begin{table}[htbp]
\footnotesize
\centering
\caption{The content details of the three real-world cavitation datasets for each flow state.}
\label{tab: CavitationDatasets-FlowStatus}
\setlength{\tabcolsep}{4mm}{
\begin{tabular}{lccccc}
\toprule 
\multirow{2}{*}{Dataset} & \multicolumn{3}{c}{Cavitation} & \multicolumn{2}{c}{Non-cavitation}  \\ 
\cmidrule{2-6} 
& \multicolumn{1}{l}{choked flow} & \multicolumn{1}{l}{constant} & \multicolumn{1}{l}{incipient} & \multicolumn{1}{l}{turbulent} & \multicolumn{1}{l}{no flow} \\ 
\midrule
Cavitation-Short           & 72    & 93   & 40   & 118   & 33  \\ 
Cavitation-Long            & 148   & 396  & 64   & 183   & 15  \\ 
Cavitation-Noise           & 40    & 40   & 40   & 40    & 0  \\ 
\bottomrule
\end{tabular}}
\end{table}

\begin{table}[htbp]
\caption{Details of the training and test sets. $(\cdot )$ denotes the number after the sliding window (a window duration of \SI{0.06}{s},  an overlap ratio of 0.25 and 512 CWT scales).}
\label{tab: training and test sets}
\footnotesize
\centering
\setlength{\tabcolsep}{0.2mm}{
\begin{tabular}{lcccccccc}
\toprule
                 \multicolumn{1}{c}{\multirow{3}{*}{Dataset}} & \multicolumn{4}{c}{Training set}      & \multicolumn{4}{c}{Testing set} \\ 
\cmidrule{2-9} 
                 & \multicolumn{3}{c}{Cavitation}       
                 & \multicolumn{1}{c|}{\multirow{2}{*}{Non}}
                 & \multicolumn{3}{c}{Cavitation}     
                 & \multirow{2}{*}{Non} \\ 
\cmidrule{2-4}
\cmidrule{6-8}
                 & Choked flow & Constant   & Incipient & \multicolumn{1}{c|}{}     & Choked flow & Constant & Incipient &    \\ 
\midrule
Cavitation-Short & 58($\times$66)    & 75($\times$66)   & 32($\times$66)  & \multicolumn{1}{c|}{121($\times$66)}  & 14($\times$66)    & 18($\times$66) & 8($\times$66)    & 30($\times$66)    \\
Cavitation-Long  & 117($\times$555)   & 315($\times$555) & 55($\times$555)  & \multicolumn{1}{c|}{158($\times$555)}  & 31($\times$555)    & 81($\times$555) & 9($\times$555)  & 40($\times$555)    \\
Cavitation-Noise & 32($\times$555)    & 32($\times$555)   & 32($\times$555)  & \multicolumn{1}{c|}{32($\times$555)}    & 8($\times$555)      & 8($\times$555)   & 8($\times$555)    & 8($\times$555)      \\ 
\bottomrule
\end{tabular}
}
\end{table}

\suppsubsection{Experimental Settings and Reproducibility}
\label{sec: appExperimental Settings and Reproducibility}

\noindent\textbf{Computing environment.} All experiments are implemented in Python 3.10.19 using PyTorch 2.8.0 with CUDA 12.9 and cuDNN 9.10.2 on an x86-64 Linux workstation. The workstation is equipped with an Intel Core i9-14900K processor, 32 logical CPU cores and two NVIDIA GeForce RTX 5090 GPUs, each with approximately 32~GB of memory. Each training job uses a single GPU, and four data-loading workers are used unless otherwise specified. The NVIDIA driver version is 595.71.05.

\noindent\textbf{Data partitioning and evaluation protocol.} A uniform training-test partition is used for all four datasets, with $80.00\%$ of the original samples assigned to the training set and the remaining $20.00\%$ assigned to the held-out test set. For the cavitation datasets, the partition is completed before sliding-window augmentation and time-frequency transformation, preventing overlapping subsequences from the same original measurement from being distributed across the training and test sets. The resulting cavitation dataset-specific sample counts are reported in Supplementary Table \ref{tab: CavitationDatasets-FlowStatus} and Supplementary Table \ref{tab: training and test sets}. A separate validation split is not used, and all models are trained for a fixed schedule of 30 epochs without early stopping. For VTPFR, 758 images are assigned to the training set and 189 images are assigned to the held-out test set. The reported classification metrics are calculated on the predefined held-out test splits after the training schedule is completed.

\noindent\textbf{Optimization and model initialization.} All deep-learning models are trained from scratch without external pre-trained weights. Cross-entropy loss is used for supervised optimization, and model parameters are updated using Adam with $\beta_1=0.9$, $\beta_2=0.999$, $\epsilon=10^{-8}$, zero weight decay and AMSGrad disabled. The initial learning rate is set to $1\times10^{-4}$, and the batch size is set to 16. A cosine learning-rate schedule $\eta_e=\eta_0\frac{1+\cos(\pi e/E)}{2}$ is used, where $\eta_0$ denotes the initial learning rate, $e$ is the current epoch and $E$ is the total number of epochs. The learning rate is updated every five epochs. To stabilize optimization, the $\ell_2$ norm of the gradients is clipped to a maximum value of 4.0. Parameters are initialized according to the original implementation of each architecture. Convolutional layers in the ResNet and DenseNet families use Kaiming initialization, whereas linear and embedding layers in Transformer-style architectures generally use a truncated normal distribution with a standard deviation of 0.02. MulT projection layers use Xavier initialization, and Perceiver-style modules follow their PyTorch default initialization when no architecture-specific rule is defined. The dropout configuration of each baseline follows its corresponding implementation rather than being overridden by a common global value.

\noindent\textbf{Input processing.} All image inputs are loaded in RGB format. DenseNet models use an input resolution of $256\times256$, whereas all other deep-learning models use $224\times224$. For the cavitation datasets, the training images are sequentially transformed using horizontal flipping with a probability of 1.0, vertical flipping with a probability of 1.0 and a fixed rotation of $180^\circ$, followed by resizing to the model-specific input resolution. During testing, no flipping or rotation is applied, and the images are only resized to the corresponding input resolution. Image intensities are subsequently converted from integer pixel values in $[0,255]$ to floating-point tensors in $[0,1]$ and normalized channel-wise as $\mathbf{x}_{c}^{\mathrm{norm}}=(\mathbf{x}_{c}-\mu_c)/\sigma_c$. To avoid information leakage, the normalization statistics are calculated using only the corresponding training split and are subsequently applied unchanged to both the training and test images. For the cavitation datasets, the channel-wise training-set mean and standard deviation are $\boldsymbol{\mu}_{\mathrm{cav}}=[0.6317,0.5517,0.2756]$ and $\boldsymbol{\sigma}_{\mathrm{cav}}=[0.2312,0.2368,0.2219]$, respectively. For VTPFR, the corresponding values are $\boldsymbol{\mu}_{\mathrm{VTPFR}}=[0.1705,0.1686,0.1539]$ and $\boldsymbol{\sigma}_{\mathrm{VTPFR}}=[0.2261,0.2222,0.2151]$. For the cavitation datasets, the acoustic waveform is first divided into fixed-duration subsequences using a sliding window. The default setting uses a window duration of \SI{0.06}{s} and an overlap ratio of 0.25, such that the window size and step size are calculated from the effective sampling frequency as $W_s=\operatorname{round}(0.06f_s)$ and $S=\operatorname{round}[W_s(1-0.25)]$, respectively. Each subsequence is transformed into a time--frequency representation using a continuous wavelet transform. A real-valued Morlet wavelet is used as the mother wavelet, and the transform is computed using an FFT-based implementation with a sampling period of $1/f_s$. Under the default setting, the scale sequence is defined as $a\in\{1,2,\ldots,512\}$. These scales are converted into the corresponding pseudo-frequencies according to the specified sampling period. The full pseudo-frequency range returned by the transform is retained, and no additional lower- or upper-frequency truncation is applied. The complex wavelet coefficients $C(a,b)$ are converted into a logarithmic amplitude representation according to $A_{\mathrm{dB}}(a,b)=20\log_{10}\left(|C(a,b)|+10^{-8}\right)$. For each subsequence, the lower and upper colour limits are set independently to the 5th and 95th percentiles of $A_{\mathrm{dB}}$, respectively. Values outside this interval are saturated at the corresponding colour limits. The resulting representation is rendered using the jet colour map, with the time axis spanning the complete window duration and the frequency axis spanning the pseudo-frequency range returned by the transform. Axes, tick labels and surrounding margins are removed, and each spectrogram is saved as an axis-free square PNG image using a $4\times4$-inch canvas at 300~dpi. The generated images are subsequently resized to the model-specific input resolution during data loading. The default settings are used for the main comparisons, whereas the window duration, overlap ratio and number of wavelet scales are varied individually in the corresponding robustness experiments. Physical variables are read from the corresponding files, converted to 32-bit floating-point values and passed directly to the multimodal models without additional normalization in the data loader. The image-only baselines receive only the response representation, whereas PhyNet and the hybrid baselines receive both the response representation and the corresponding physical variables.

\noindent\textbf{PhyNet configuration.} The default PhyNet configuration uses three input channels and divides each $224\times224$ response image into non-overlapping $16\times16$ patches. The embedding dimension is 512, the number of attention heads is 8 and the encoder contains 3 stacked layers. Each feed-forward block has a hidden dimension of 2048 and uses the GELU activation function. A dropout rate of 0.1 is applied in PhyNet. The model produces four output classes for each benchmark. The same default configuration is retained unless the corresponding architectural factor is explicitly varied in the sensitivity experiments.

\noindent\textbf{Gradient-based localization analysis.} Gradient-based localization maps are generated using Grad-CAM to examine how physical-variable conditioning affects the local response evidence used for prediction. For the full PhyNet and the variant without physical-to-response cross-attention, the gradient of the predicted-class score is computed with respect to the output features of the image-embedding projection layer. These gradients are spatially averaged to obtain channel-wise importance weights, which are then used to form a weighted combination of the corresponding feature maps. A rectified linear unit is applied to retain only positive contributions to the predicted class. The resulting patch-level localization map has a spatial resolution of $16\times16$. Each map is independently min-max normalized to the range $[0,1]$, upsampled to the input resolution of $224\times224$, and lightly smoothed to reduce patch-boundary artefacts. The processed localization map is then overlaid semi-transparently on a desaturated version of the original time--frequency spectrogram. To ensure a consistent comparison, the full and ablated models are evaluated using the same input samples, target classes, localization procedure and visualization settings, including a unified positive colour scale. For the VTPFR analysis, the same procedure and visualization settings are applied to both PhyNet and PerceiverIO.

\noindent\textbf{Masking and robustness experiments.} Physical-variable masking is used only for post-training interpretation and ablation analyses and is not applied during model training. When a variable is masked, its value is replaced by the empirical mean of that variable calculated from the corresponding dataset; zero replacement and learnable mask tokens are not used. Single-variable masking removes one variable at a time, whereas combination masking evaluates subsets containing different numbers of retained or masked variables. Sampling-frequency robustness on Cavitation-Short is evaluated using the original sampling frequency of \SI{1562500}{Hz} and nominal target frequencies of \SI{781250}{Hz}, \SI{390625}{Hz}, \SI{260416}{Hz}, \SI{195312}{Hz} and \SI{48000}{Hz}. Resampling is applied to each complete acoustic waveform before sliding-window segmentation and CWT generation. For the evaluated sampling frequencies, the signals are resampled using a polyphase resampling procedure. Integer decimation ratios are used when the target frequency is sufficiently close to an integer fraction of the original frequency; otherwise, the target-to-original sampling-rate ratio is represented as a rational upsampling and downsampling pair. The polyphase resampling procedure incorporates low-pass anti-alias filtering. The effective sampling frequency is calculated from the applied rational ratio and recorded for each generated sample. After resampling, the window size and step size are recalculated from the effective sampling frequency to retain the default window duration of \SI{0.06}{s} and overlap ratio of 0.25. The same Morlet wavelet, scale sequence and image-generation procedure are then applied at each sampling frequency. Noise robustness is evaluated at signal-to-noise ratios of $-5$, 0, 5, 10, 20 and 30~dB. For a clean acoustic signal $\mathbf{x}=\{x_i\}_{i=1}^{N}$, the signal power is calculated as $P_{\mathrm{s}}=\frac{1}{N}\sum_{i=1}^{N}|x_i|^2$. For a specified signal-to-noise ratio $\mathrm{SNR}_{\mathrm{dB}}$, the target noise power and standard deviation are defined as $P_{\mathrm{n}}=\frac{P_{\mathrm{s}}}{10^{\mathrm{SNR}_{\mathrm{dB}}/10}}$ and $\sigma_{\mathrm{n}}=\sqrt{P_{\mathrm{n}}}$, respectively. Independent zero-mean Gaussian noise is then sampled as $\boldsymbol{\epsilon}\sim\mathcal{N}(\mathbf{0},\sigma_{\mathrm{n}}^2\mathbf{I})$, and the noisy signal is generated as $\widetilde{\mathbf{x}}=\mathbf{x}+\boldsymbol{\epsilon}$. Noise is added to the complete acoustic waveform before sliding-window segmentation and CWT generation. A separate deterministic random seed is constructed for each file--SNR pair as $s_{\mathrm{noise}}=\left[n_{\mathrm{file}}\times1000003+\operatorname{round}\bigl((\mathrm{SNR}_{\mathrm{dB}}+100)\times100\bigr)\right]\bmod 2^{32}$, where $n_{\mathrm{file}}$ denotes the numerical identifier of the original signal file. Consequently, different SNR levels use different noise realizations for the same signal, while repeated preprocessing with the same file and SNR reproduces the same noisy waveform. When the clean signal power is zero, the original signal is retained unchanged. Cross-duration transfer is assessed in both directions between Cavitation-Short and Cavitation-Long. For each robustness experiment, only the factor under investigation is varied, while the remaining data partitioning, preprocessing, architecture and optimization settings are fixed at their default values.

\noindent\textbf{Asymmetric fusion ablation.} To examine the contribution of each component in the asymmetric fusion design, five ablation variants are constructed on Cavitation-Long. The \emph{w/o physical variables} variant removes the entire physical-variable branch and retains only the response self-attention pathway. The \emph{w/o global response token} variant removes the learnable global response token and allows physical-variable tokens to interact directly with local response tokens. In the \emph{w/o cross-attention} variant, the physical-to-response cross-attention module is replaced by a simple residual injection block, defined as ${\rm{Concat}}\left[global\_cls,\,pooled\_physical\right]\rightarrow{\rm{Linear}}(\cdot)$, while the global response token is retained. The \emph{w/o self-attention propagation} variant preserves the physical-to-response cross-attention module and the global response token but removes the subsequent propagation of global constraints back to local response tokens through response self-attention. The \emph{naive fusion} variant removes the asymmetric conditioning mechanism from the encoder. In this variant, the response branch is encoded using only self-attention, whereas the physical-variable branch is embedded and pooled independently. The classifier then applies ${\rm{Concat}}\left[global\_cls,\,pooled\_patch,\,pooled\_physical\right]\rightarrow{\rm{Linear}}(\cdot)$ without cross-attention or token-level conditional injection. All ablation variants use the same data partition, input preprocessing, optimization settings and training schedule as the complete PhyNet model.

\suppsubsection{Evaluation Metrics}
\label{sec: Evaluation Metrics}

For a classification task with $K$ classes, let $\mathrm{TP}_k$, $\mathrm{FP}_k$, $\mathrm{TN}_k$ and $\mathrm{FN}_k$ denote the numbers of true positives, false positives, true negatives and false negatives for class $k$ under a one-vs-rest formulation, respectively. The support of class $k$ is defined as $n_k=\mathrm{TP}_k+\mathrm{FN}_k$, and the total number of samples is $N=\sum_{k=1}^{K}n_k$. The overall classification accuracy is calculated as:
\begin{equation}
\mathrm{Accuracy}=\frac{\sum_{k=1}^{K}\mathrm{TP}_k}{N}.
\end{equation}

The class-specific precision, recall and F1-score are defined as:
\begin{equation}
\begin{aligned}
\mathrm{Precision}_k&=\frac{\mathrm{TP}_k}{\mathrm{TP}_k+\mathrm{FP}_k},\\
\mathrm{Recall}_k&=\frac{\mathrm{TP}_k}{\mathrm{TP}_k+\mathrm{FN}_k},\\
\mathrm{F1}_k&=\frac{2\,\mathrm{Precision}_k\,\mathrm{Recall}_k}{\mathrm{Precision}_k+\mathrm{Recall}_k}.
\end{aligned}
\end{equation}

To account for differences in class frequency, precision, recall and F1-score are reported as support-weighted averages:
\begin{equation}
\begin{aligned}
\mathrm{Precision}_{\mathrm{w}}&=\sum_{k=1}^{K}\frac{n_k}{N}\,\mathrm{Precision}_k,\\
\mathrm{Recall}_{\mathrm{w}}&=\sum_{k=1}^{K}\frac{n_k}{N}\,\mathrm{Recall}_k,\\
\mathrm{F1}_{\mathrm{w}}&=\sum_{k=1}^{K}\frac{n_k}{N}\,\mathrm{F1}_k.
\end{aligned}
\end{equation}

Receiver operating characteristic curves are calculated using a one-vs-rest strategy. For each class, the true positive rate and false positive rate at decision threshold $\tau$ are defined as:
\begin{equation}
\begin{aligned}
\mathrm{TPR}_k(\tau)&=\frac{\mathrm{TP}_k(\tau)}{\mathrm{TP}_k(\tau)+\mathrm{FN}_k(\tau)},\\
\mathrm{FPR}_k(\tau)&=\frac{\mathrm{FP}_k(\tau)}{\mathrm{FP}_k(\tau)+\mathrm{TN}_k(\tau)}.
\end{aligned}
\end{equation}

The micro-average ROC curve is obtained by aggregating the one-vs-rest decisions across all classes before calculating the true positive and false positive rates:
\begin{equation}
\begin{aligned}
\mathrm{TPR}_{\mathrm{micro}}(\tau)&=\frac{\sum_{k=1}^{K}\mathrm{TP}_k(\tau)}{\sum_{k=1}^{K}\left[\mathrm{TP}_k(\tau)+\mathrm{FN}_k(\tau)\right]},\\
\mathrm{FPR}_{\mathrm{micro}}(\tau)&=\frac{\sum_{k=1}^{K}\mathrm{FP}_k(\tau)}{\sum_{k=1}^{K}\left[\mathrm{FP}_k(\tau)+\mathrm{TN}_k(\tau)\right]}.
\end{aligned}
\end{equation}

The micro-average area under the ROC curve is defined as:
\begin{equation}
\mathrm{AUC}_{\mathrm{micro}}=\int_{0}^{1}\mathrm{TPR}_{\mathrm{micro}}\bigl(\mathrm{FPR}_{\mathrm{micro}}\bigr)\,d\bigl(\mathrm{FPR}_{\mathrm{micro}}\bigr).
\end{equation}

In practice, the class-specific and micro-average AUC values are numerically approximated using the trapezoidal rule over the discrete points of the corresponding ROC curves.

\suppsection{Supplementary Results}
\label{sec: Results}
\suppsubsection{Class-wise error patterns and ROC analysis}
\label{subsec: Class-wise error patterns and ROC analysis}
Additional confusion matrices and ROC curves for the remaining datasets (Cavitation-Short and Cavitation-Noise) are provided in this section, as shown in Supplementary Figure \ref{fig: AppConfusionROC}. These supplementary results complement the main text by extending the class-wise error analysis and threshold-independent performance analysis.
\begin{figure*}[htbp]
    \centering
    \includegraphics[width=\textwidth,height=105mm]{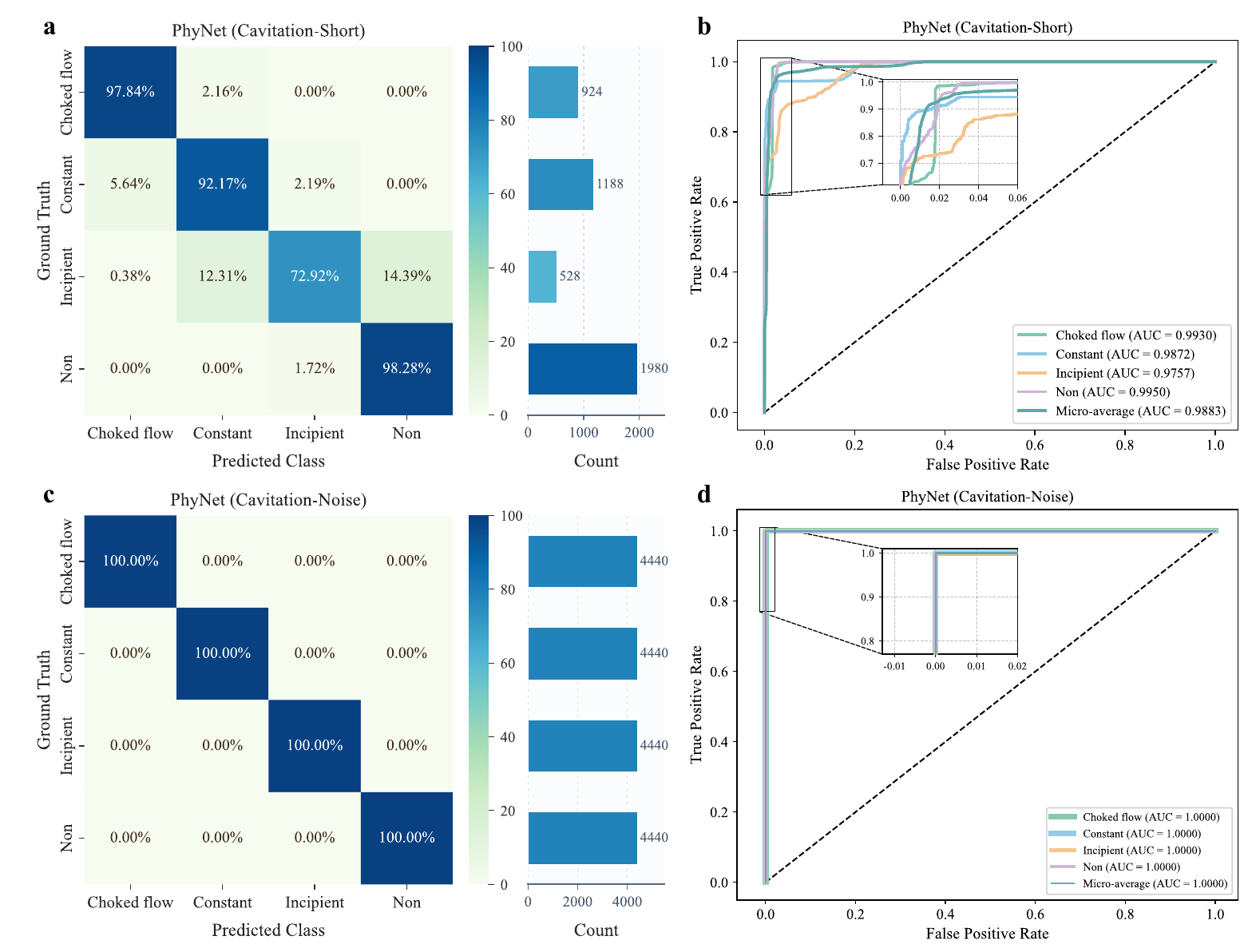}
    \caption{\textbf{Class-wise error patterns and ROC analysis on Cavitation-Short and Cavitation-Noise datasets}. \textbf{a, c}, Confusion matrices of PhyNet on Cavitation-Short and Cavitation-Noise, respectively. \textbf{b, d}, Receiver operating characteristic (ROC) curves of PhyNet on Cavitation-Short and Cavitation-Noise, respectively. The confusion matrices visualize class-wise prediction patterns and misclassification at the chosen decision threshold, whereas the ROC curves illustrate the relationship between the true positive rate and the false positive rate across thresholds. The area under each ROC curve corresponds to the AUC value, which provides a threshold-independent measure of discriminative performance.}
    \label{fig: AppConfusionROC}
\end{figure*}

\suppsubsection{Condition-aware attention}
\label{subsec: Condition-aware attention}
\noindent\textbf{Head-Wise Cross-Attention Maps.} This part further details the head-wise physical-variable attention patterns, as shown in Supplementary Figure \ref{fig: AppHeadCrossAttention}. By presenting head-wise cross-attention maps for representative high-confidence prototype samples and low-margin boundary samples across the four flow states, we show that the dominant contributions of vapour pressure $p_v$, volumetric flow rate $Q$ and pressure recovery factor ${X_f}$ emerge from structured head specialization rather than from homogeneous attention allocation.
\begin{figure*}[htbp]
    \centering
    \includegraphics[width=\textwidth,height=140mm]{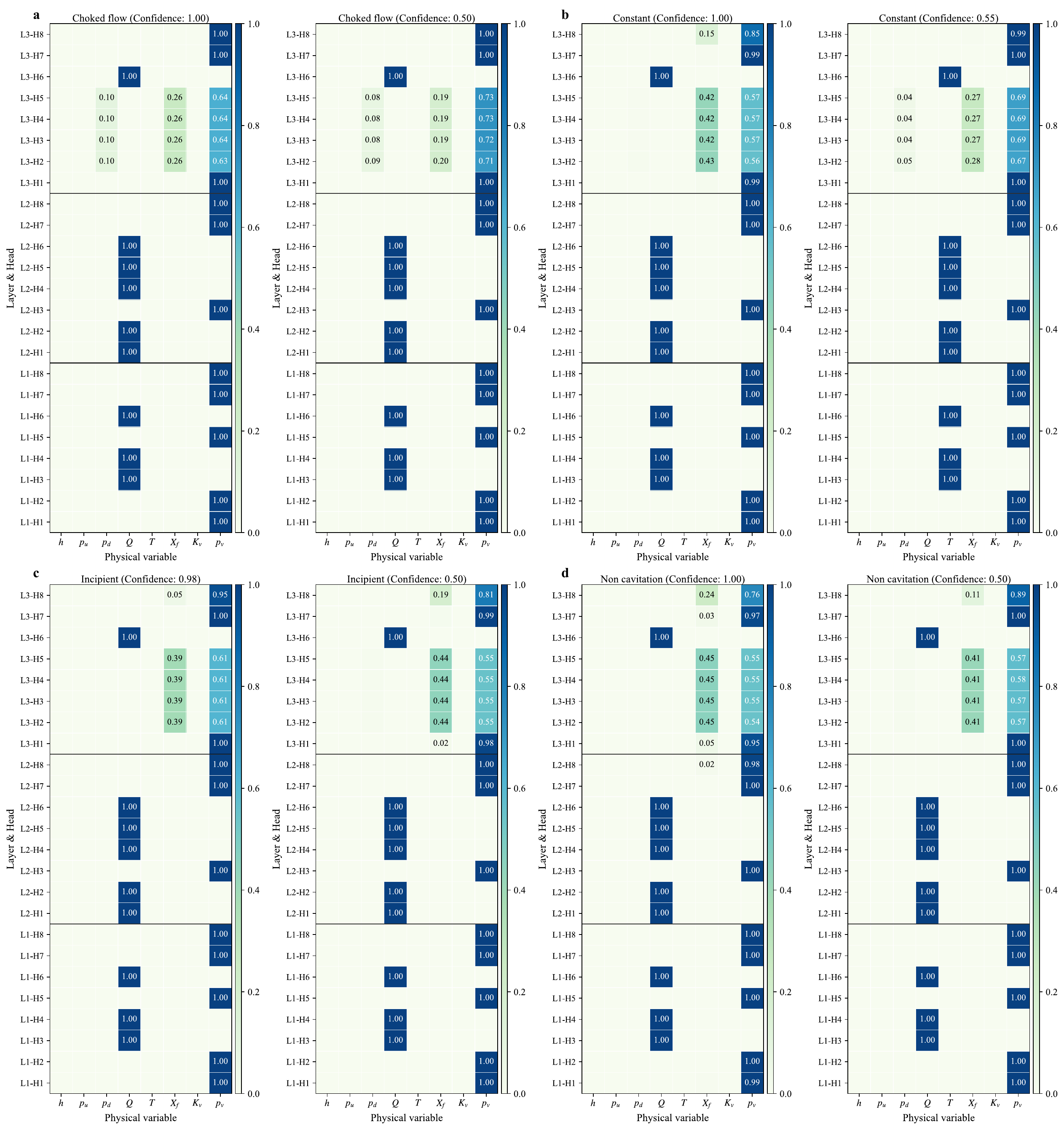}
    \caption{\textbf{Head-wise physical-variable attention patterns on Cavitation-Long}. \textbf{a-d}, Head-wise cross-attention weights from the global response token to physical-variable tokens for representative samples from choked-flow cavitation (a), constant cavitation (b), incipient cavitation (c) and non-cavitation (d). For each state, the left panel shows a high-confidence prototype sample and the right panel shows a low-margin boundary sample. Rows correspond to different attention heads across the three cross-attention layers and columns correspond to the physical variables. The visualizations show that the class-level patterns arise from structured head specialization rather than uniform weighting across heads. In particular, most heads assign dominant weights to vapour pressure $p_v$, while a smaller subset selectively attends to volumetric flow rate $Q$ and pressure recovery factor ${X_f}$.}
    \label{fig: AppHeadCrossAttention}
\end{figure*}

\noindent\textbf{Layer-Wise Mean Cross-Attention Weights on VTPFR.} 
To further evaluate how physical variables modulate visual response representations in two-phase-flow recognition, we analyse the layer-wise mean cross-attention weights learned by the physical-to-response cross-attention module on VTPFR, as shown in Supplementary Figure \ref{fig: CrossAttention_VTPFR}. Representative high-confidence prototype samples and low-margin boundary samples from Taylor bubble, slug, churn and bubbly regimes show that PhyNet consistently assigns the largest attention weights to kick influx pressure $p_k$. In addition, the gas type $g_t$, injection time $t_i$ and the selected flow-related variables also make additional contributions, with their relative importance varying across regimes. This core pattern remains visible even in boundary samples, indicating that the model relies on a stable set of physically meaningful operating variables when separating visually adjacent and difficult flow regimes, as shown in Supplementary Figure \ref{fig: CrossAttention_VTPFR}a-\ref{fig: CrossAttention_VTPFR}b. The class-averaged statistics in Supplementary Figure \ref{fig: CrossAttention_VTPFR}c further confirm that kick influx pressure $p_k$ dominates across all four regimes, whereas the secondary contributions of gas type $g_t$, injection time $t_i$ and other operational variables are more regime-dependent. 

This pattern is physically plausible for vertical gas-liquid two-phase flow. Operating pressure directly influences gas density, bubble expansion, void-fraction evolution and interfacial dynamics, all of which affect flow-regime formation and transition. Likewise, the injected gas condition and its duration are closely related to gas accumulation, interface organization and the temporal development of Taylor-bubble, slug, churn and bubbly structures. In general, gas-liquid flow regimes in vertical annuli are known to depend strongly on operating pressure and on the balance between gas and liquid transport conditions. The observation that kick influx pressure $p_k$ plays a dominant role, whilst gas type $g_t$, injection time $t_i$ and other operational variables provide regime-dependent secondary contributions is consistent with established physical principles of two-phase flow rather than reflecting an arbitrary attention pattern.

To determine whether these attention patterns are functionally relevant, we further relate them to recognition performance under physical-variable masking. As shown in Supplementary Figure \ref{fig: CrossAttention_VTPFR}d, the accuracy declines progressively as larger numbers of physical variables are randomly masked, indicating that PhyNet depends on the joint operating context rather than on any single variable alone. The whole-test-set importance statistics in Supplementary Figure \ref{fig: CrossAttention_VTPFR}e again highlight kick influx pressure $p_k$ as the most influential variable overall, while gas type $g_t$, injection time $t_i$ and selected flow-related variables provide smaller but non-negligible contributions. At the same time, the single-variable masking results in Supplementary Figure \ref{fig: CrossAttention_VTPFR}f suggest that the functional contribution of each variable to the final prediction does not necessarily follow exactly the same ordering as the average cross-attention weights. Instead, the attention maps and masking experiments provide complementary views of variable importance: the former highlight the most consistently attended operating cues, whereas the latter reflect how strongly each variable supports the final decision once interactions and partial redundancy among variables are taken into account. Together, these results indicate that PhyNet relies on a structured set of physically meaningful operating constraints, with $p_k$ serving as the most stable dominant variable and several regime-dependent secondary variables jointly supporting robust recognition in heterogeneous two-phase flow conditions.

The head-wise visualizations in Supplementary Figure \ref{fig: AppHeadCrossAttention_VTPFR} further show that these regime-level attention patterns arise from structured head specialization rather than uniform weighting across heads. Across different regimes, multiple heads consistently emphasize $p_k$, whereas other heads selectively respond to $g_t$, $t_i$ and several flow-related variables depending on regime type and decision difficulty. This suggests that the model does not compress all physical constraints into a single homogeneous weighting pattern, but instead distributes complementary operating cues across different attention heads. Such structured specialization provides a plausible mechanism by which PhyNet preserves both a stable dominant pressure cue and regime-specific secondary constraints in its cross-modal interaction.
\begin{figure*}[htbp]
    \centering
    \includegraphics[width=\textwidth,height=105mm]{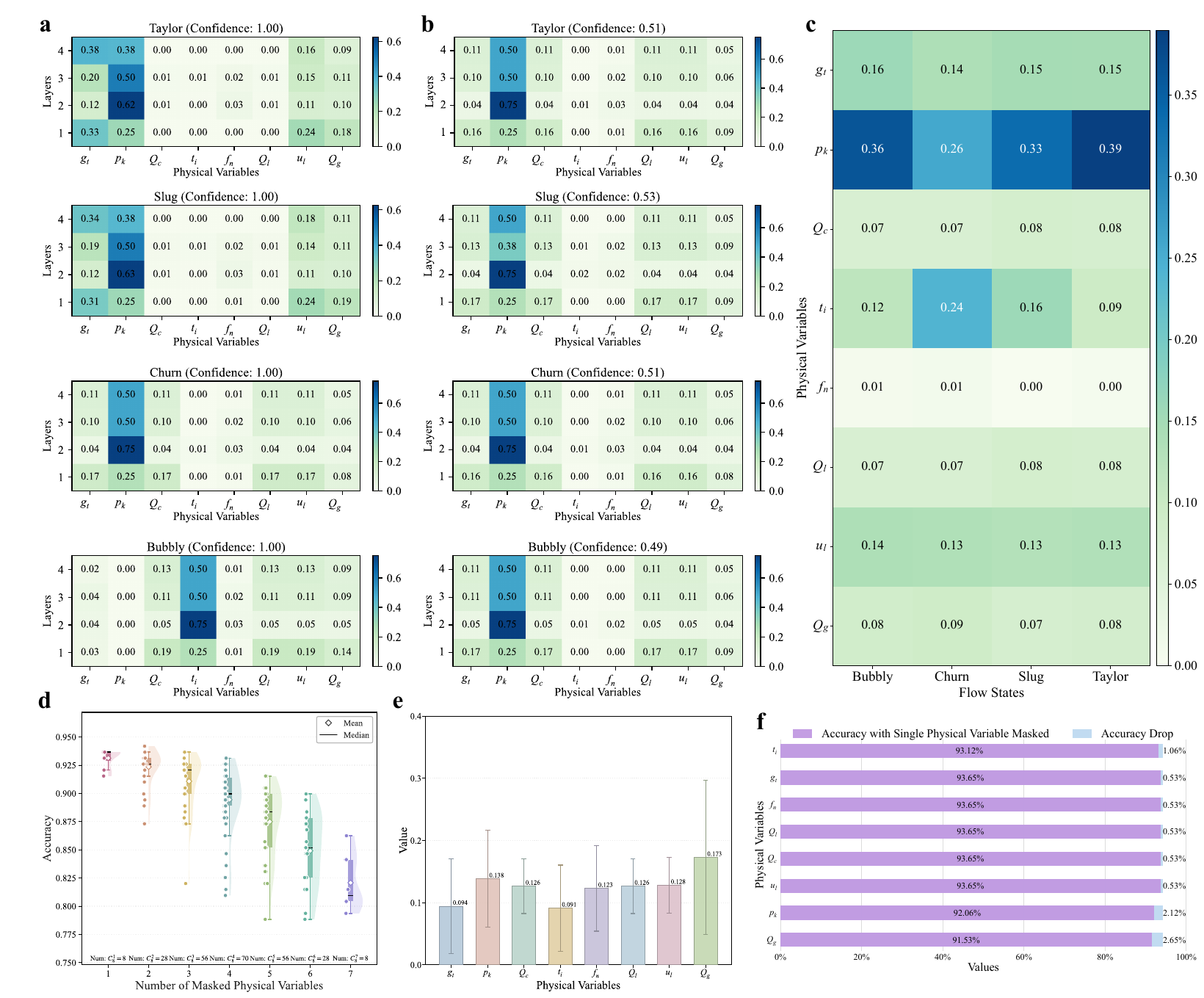}
    \caption{\textbf{Physical-response cross-attention and its association with recognition performance on VTPFR}. \textbf{a}, \textbf{b}, Layer-wise mean cross-attention weights from the global response token to physical-variable tokens for representative high-confidence prototype samples (a) and low-margin boundary samples (b) from the four regimes. High-confidence prototype samples denote correctly classified examples with high prediction confidence, whereas low-margin boundary samples denote correctly classified examples with small margins between the top two predicted classes. \textbf{c}, Class-averaged physical-variable attention across bubbly, churn, slug and Taylor regimes. \textbf{d}, Accuracy under randomly masked physical-variable combinations. Random masking of different numbers of physical variables is equivalent to sampling variable subsets of different sizes. \textbf{e}, Physical-variable importance across the whole test set of VTPFR. \textbf{f}, Accuracy under single-variable masking. The accuracy drop is defined relative to the full-model performance of PhyNet on VTPFR. The reported values are averaged over three independent model-training runs ($n=3$), and the error bars denote one standard deviation. Across representative samples and class-averaged statistics, kick influx pressure $p_k$ consistently receives the largest attention weights, while gas type $g_t$, injection time $t_i$ and selected flow-related variables provide additional regime-dependent contributions. The masking analyses further show that recognition performance depends on a structured set of operating variables rather than on any single auxiliary cue in isolation. Together, the attention and masking results provide complementary views of variable importance, indicating that PhyNet captures physically meaningful operating constraints for robust recognition in heterogeneous two-phase-flow conditions.}
    \label{fig: CrossAttention_VTPFR}
\end{figure*}
\begin{figure*}[htbp]
    \centering
    \includegraphics[width=\textwidth,height=140mm]{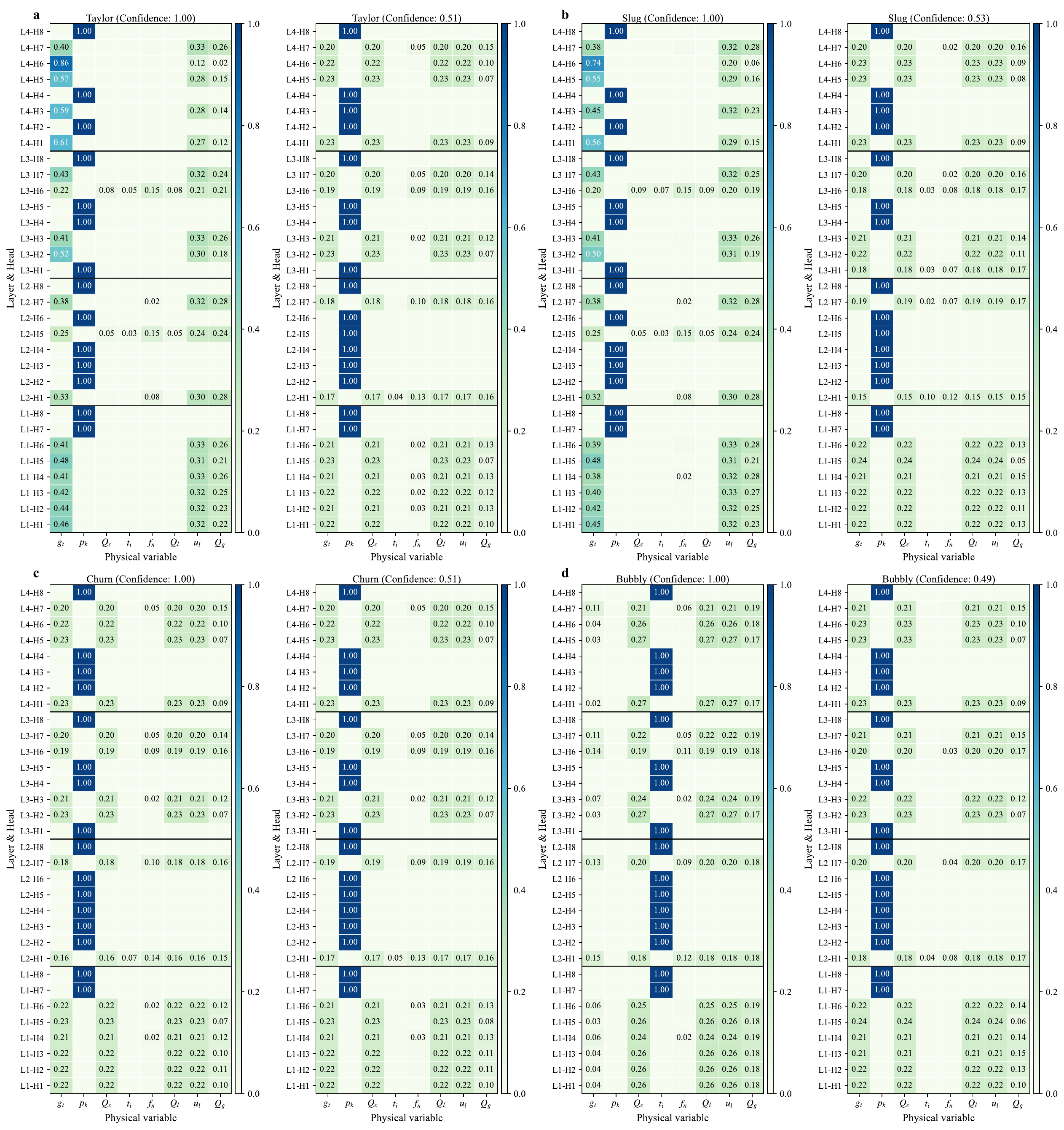}
    \caption{\textbf{Head-wise physical-variable attention patterns on VTPFR}. \textbf{a-d}, Head-wise cross-attention weights from the global response token to physical-variable tokens for representative samples from Taylor bubble (a), slug (b), churn (c) and bubbly (d) regimes. For each regime, the left panel shows a high-confidence prototype sample and the right panel shows a low-margin boundary sample. Rows correspond to different attention heads across the four cross-attention layers and columns correspond to the physical variables. The visualizations show that the regime-level attention patterns arise from structured head specialization rather than uniform weighting across heads.}
    \label{fig: AppHeadCrossAttention_VTPFR}
\end{figure*}

\noindent\textbf{Gradient-based Localization Maps on VTPFR.} To further evaluate whether physical-variable conditioning also reshapes local visual evidence in two-phase-flow recognition, we visualize gradient-based localization maps on representative VTPFR samples, as shown in Supplementary Figure \ref{fig: AppSelfAttention_VTPFR}. Across Taylor bubble, slug, churn and bubbly regimes, PhyNet generally produces more concentrated and structurally meaningful response regions  than the strongest hybrid baseline, PerceiverIO. In particular, the highlighted regions in PhyNet are more consistently aligned with the salient gas-liquid interface structures and regime-specific flow patterns, whereas PerceiverIO often exhibits more diffuse or fragmented evidence allocation. These observations suggest that PhyNet improves global class discrimination while physical-variable conditioning helps guide the model towards more relevant local visual evidence in heterogeneous flow-regime recognition.
\begin{figure*}[htbp]
    \centering
    \includegraphics[width=\textwidth,height=155mm]{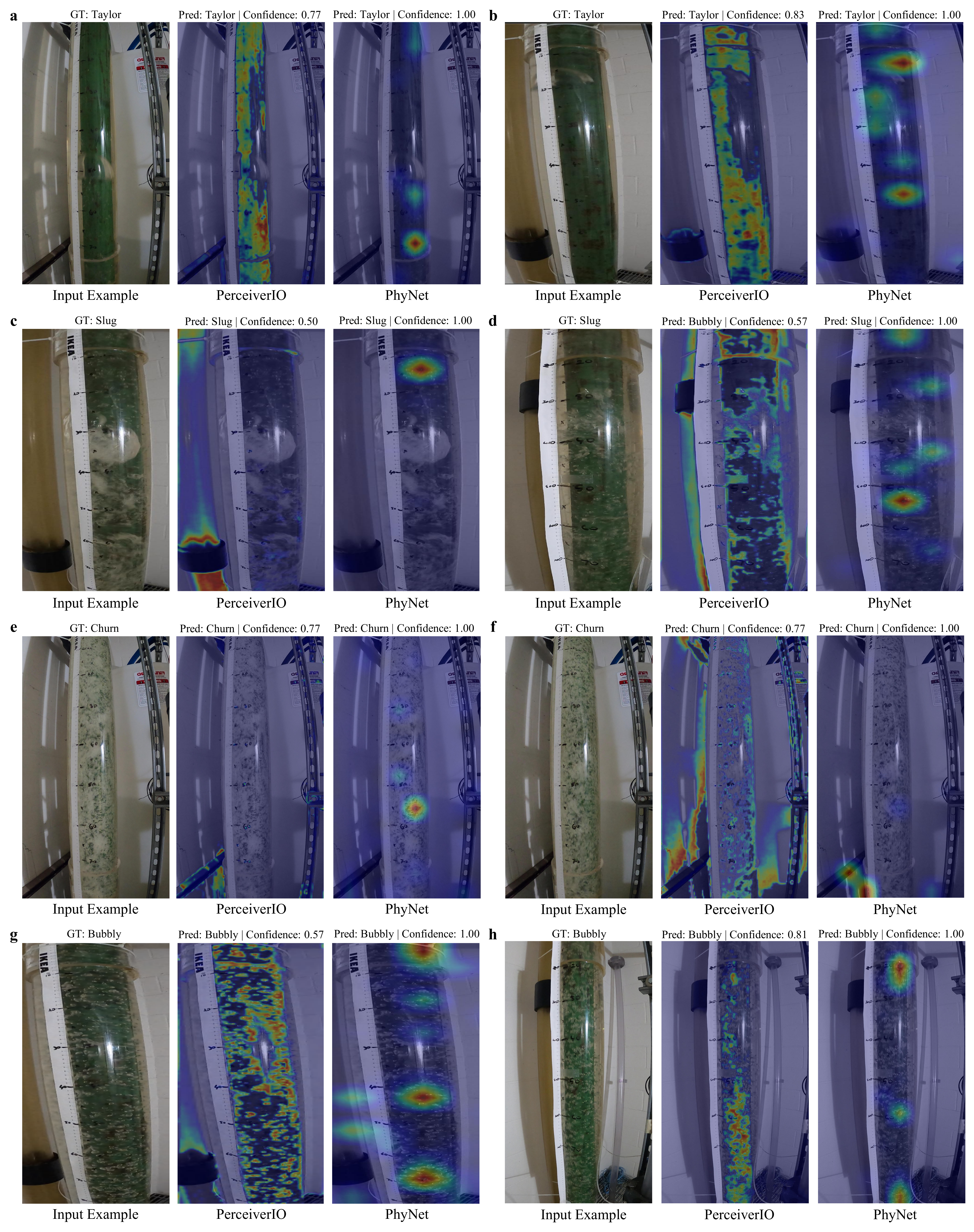}
    \caption{\textbf{Local response evidence under physical variables on VTPFR}. \textbf{a-h}, Representative annular-flow images from the four flow regimes, including Taylor bubble (a,b), slug (c,d), churn (e,f) and bubbly (g,h). For each example, the left panel shows the input image, the middle panel shows the gradient-based localization overlay of PerceiverIO, and the right panel shows the corresponding localization overlay of PhyNet. The localization maps are generated by Grad-CAM from the image-embedding projection layer. Compared with PerceiverIO, PhyNet produces more compact and physically relevant localized evidence that is better aligned with the characteristic flow structures of each regime, while PerceiverIO tends to yield more diffuse, fragmented or less discriminative response patterns. The confidence scores shown above each panel correspond to the predicted class probabilities of the respective models.}
    \label{fig: AppSelfAttention_VTPFR}
\end{figure*}

\suppsubsection{Robustness analysis}
\label{subsec: Robustness analysis}
This section provides the robustness results for the remaining evaluation metrics, as shown in Supplementary Figure \ref{fig: AppRobustnessAnalysis}. In parallel with the accuracy-based analysis, we report precision, recall and F1-score under the same perturbation settings, including model hyperparameters, acquisition quality and cross-duration transfer. These supplementary results extend the robustness evaluation and confirm that the performance stability of PhyNet is not limited to a single metric.
\begin{figure*}[htbp]
    \centering
    \includegraphics[width=0.8\textwidth,height=155mm]{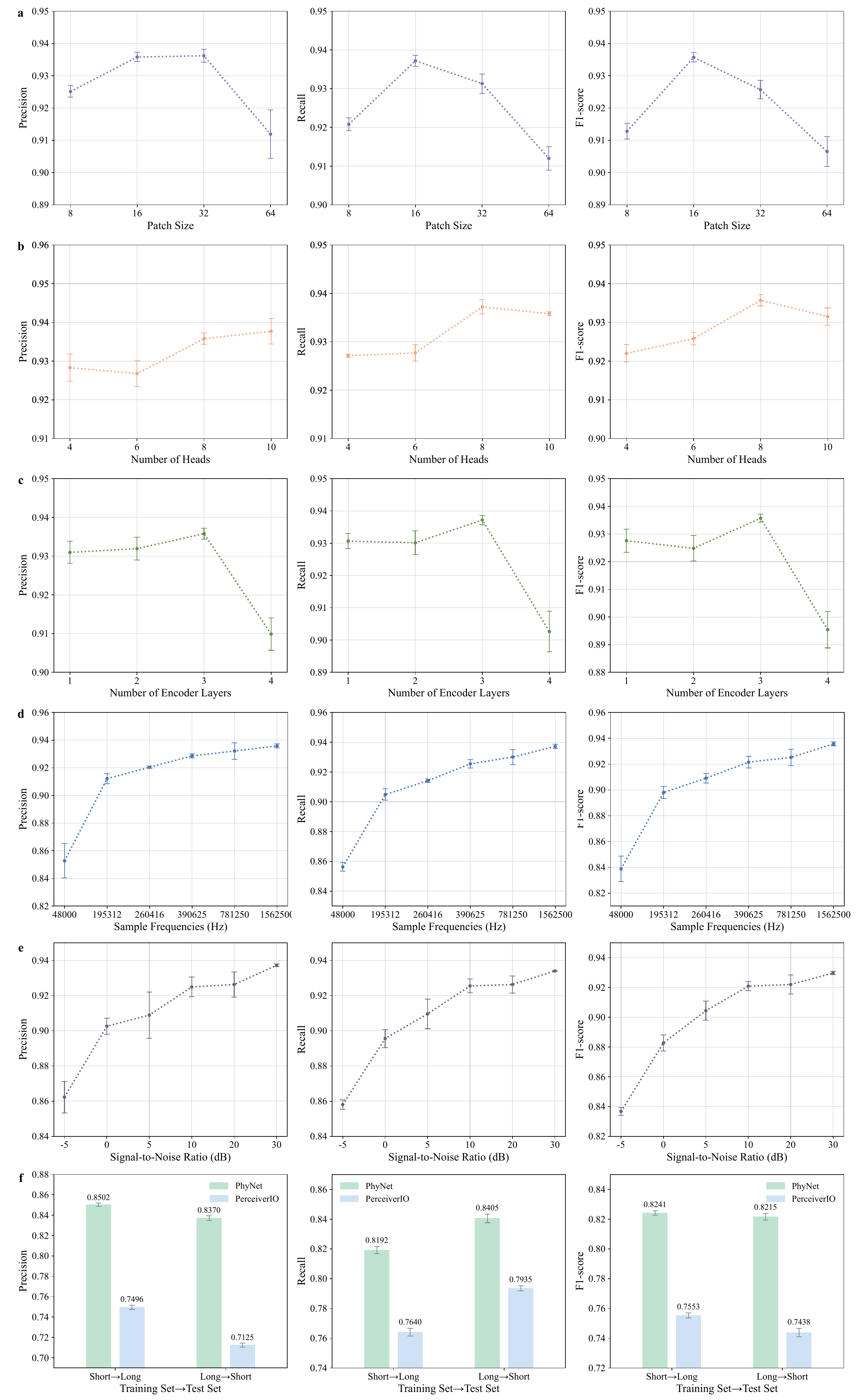}
    \caption{\textbf{Robustness results of PhyNet for the remaining evaluation metrics}. \textbf{a-c}, Sensitivity of PhyNet to architectural hyperparameters on Cavitation-Short, including patch size (a), number of attention heads (b) and number of encoder layers (c), reported in terms of Precision, Recall and F1-score. \textbf{d-e}, Precision, Recall and F1-score of PhyNet across different sampling frequencies (d) and different signal-to-noise ratios (e). \textbf{f}, Performance comparison between PhyNet and PerceiverIO during cross-duration transfer between Cavitation-Short and Cavitation-Long (Training set$\to$Test set: Short$\to$Long and Long$\to$Short), reported in terms of Precision, Recall and F1-score. The reported values are averaged over three independent model-training runs ($n=3$), and the error bars denote one standard deviation.}
    \label{fig: AppRobustnessAnalysis}
\end{figure*}

\else

\title[Physical-Variable-Informed Deep Learning for Robust Diagnosis in Industrial Fluid Transport
Systems]{Physical-Variable-Informed Deep Learning for Robust Diagnosis in Industrial Fluid Transport
Systems}

\author[1,4]{\fnm{Yu} \sur{Sha} \orcidlink{0000-0003-4521-2077}}
\author[1,3]{\fnm{Junqi} \sur{Tao} \orcidlink{0000-0002-4406-2284}}
\author[2]{\fnm{Dixin} \sur{Zhou}\orcidlink{0009-0006-5509-785X}} 
\author[4,5,6]{\fnm{Horst} \sur{Stoecker} \orcidlink{0000-0002-3282-3664}}
\author[7]{\fnm{Domagoj} \sur{Vnucec} \orcidlink{0000-0002-2752-9612}}
\author[7]{\fnm{Andreas} \sur{Widl} \orcidlink{0009-0009-0664-2024}}
\author*[1,2,4]{\fnm{Kai} \sur{Zhou} \orcidlink{0000-0001-9859-1758}}\email{zhoukai@cuhk.edu.cn}

\affil[1]{\orgdiv{School of Science and Engineering}, \orgname{The Chinese University of Hong Kong}, \city{Shenzhen}, \postcode{518172}, \country{China}}
\affil[2]{\orgdiv{School of Artificial Intelligence}, \orgname{The Chinese University of Hong Kong}, \city{Shenzhen}, \postcode{518172}, \country{China}}
\affil[3]{\orgdiv{Key Laboratory of Quark \& Lepton Physics and Institute of Particle Physics}, \orgname{Central China Normal University}, \city{Wuhan}, \postcode{430079}, \country{China}}
\affil[4]{\orgname{Frankfurt Institute for Advanced Studies}, \city{Frankfurt am Main}, \postcode{60438}, \country{Germany}}
\affil[5]{\orgdiv{Institut f{\"u}r Theoretische Physik}, \orgname{Goethe Universit{\"a}t Frankfurt}, \city{ Frankfurt am Main}, \postcode{60438}, \country{Germany}}
\affil[6]{\orgname{GSI Helmholtzzentrum f{\"u}r Schwerionenforschung GmbH}, \city{Darmstadt}, \postcode{64291}, \country{Germany}}
\affil[7]{\orgname{SAMSON AG}, \city{Frankfurt am Main}, \postcode{60314}, \country{Germany}}


\abstract{Industrial fluid transport systems are monitored through heterogeneous sensing streams, but robust fault diagnosis remains difficult because response signals vary with operating conditions and differ from physical variables in sampling rate, observation duration and data granularity. To address these challenges, we propose a physical-variable-informed deep learning framework (PhyNet) for robust fault diagnosis under real-world sampling heterogeneity. PhyNet treats physical variables as global operating constraints and response signals as local diagnostic evidence. Its physical-to-response cross-attention mechanism connects sequence-level physical-variable tokens with a learnable global response token, enabling interaction without strict temporal alignment. Across four benchmarks, including three real-world cavitation datasets and one public two-phase-flow dataset, PhyNet achieves the best performance across all evaluated metrics. Class-wise analyses, together with attention and masking analyses, show improved discrimination of ambiguous states and indicate that physical variables meaningfully constrain local response interpretation. These findings demonstrate a deployment-oriented approach to robust diagnosis in heterogeneous industrial environments.}




\maketitle

\section*{}
\label{sec: 1_Introduction}
Industrial fluid transport systems, including control valves, pumps and pipelines, are critical infrastructure for energy, chemical and manufacturing industries, whose safe operation underpins continuous production \cite{tan2025real}. When faults occur and persist within the system, they can exacerbate vibration, noise and erosion, potentially leading to fluid leakage and even cascading accidents \cite{xu2021flow,crum1986acoustic,yusha2022regional}. Although these systems are increasingly monitored using continuous, multi-source and high-precision sensing, reliable fine-grained diagnosis remains challenging because fault precursors are subtle and monitoring signals are strongly affected by operating-state variations \cite{kauffmann2025explainable,scheffer2009early,yusha2024hierarchical}. For instance, characteristic signals during incipient cavitation are often weak and can be easily masked by noise and operational fluctuations. The same fault intensity can also produce different response patterns as load, flow rate or temperature changes, degrading deployed models \cite{kauffmann2025explainable,geirhos2020shortcut}.

Existing approaches primarily address this challenge along three directions, but each exhibits inherent industrial limitations \cite{sha2025review}. Data-driven models typically employ convolutional neural networks (CNNs) or Transformers to learn discriminative features and patterns from response signal measurements in an end-to-end manner \cite{rabeh2025benchmarking,sha2022multi,gou2024hierarchical}. However, these methods easily latch onto condition-specific representation patterns and learn spurious correlations, resulting in weak cross-condition generalization when operating regimes change. Physics-driven methods generally utilise conventional machine-learning models to predict system states based on routinely recorded physical variables during operation \cite{naguib2025thermal,sha2022acoustic,peng2026identification,yusha2026deep}. While these methods offer clearer interpretability and are more sensitive to operational variations, they can hardly detect weak fault precursors and discriminate fine-grained intensity levels due to their reliance on macro-grained operating parameters. Hybrid multimodal models demonstrate significant potential in artificial intelligence by fusing two or more data sources \cite{fitzpatrick2023multi,borah2024synergizing}. These methods often assume temporal alignment and consistent input granularity \cite{lu2025multimodal}. In industrial monitoring scenarios, physical variables and high-frequency response signals commonly exhibit mismatched sampling rates, inconsistent sequence lengths and complex missingness patterns, making these assumptions difficult to satisfy \cite{mao2026robust,mostafavi2024wind}. Therefore, merely applying existing hybrid multimodal models can obscure critical causal cues and introduce misalignment noise, limiting their straightforward transfer and reliable deployment in industrial systems.  

High-frequency response measurements contain detailed local evidence of fault-related dynamics, but their statistical properties are strongly modulated by operating conditions, making representations that rely solely on response signals susceptible to non-stationary variations. In contrast, physical variables summarize the system operating conditions and provide physical constraints at the level of fault mechanisms, determining the physical plausibility of a fault occurrence and its manifestation in response signals. Their integration provides two complementary functions. Contextual disambiguation separates genuine fault signatures from benign fluctuations induced by changing operating conditions, while physical regularization encourages representations that remain compatible with system-level mechanisms. Both functions are particularly important for fine-grained diagnosis, where weak precursors can resemble random disturbances and where the same local response may have different meanings under different operating states. Because physical variables are often sampled sparsely or irregularly and may cover different durations, an effective framework should use them as global constraints or causal cues that modulate local time-frequency responses without one-to-one correspondence.

To address this issue, we propose a physical-variable-informed deep learning framework (PhyNet), which enables robust and resilient diagnosis of industrial fluid transport systems under real-world sampling heterogeneity. PhyNet follows an asymmetric representation principle, treating physical variables as global constraint carriers and signal responses as local diagnostic evidence. In detail, PhyNet embeds each physical variable sequence as a sequence-wise variate token, which accommodates differences in sampling frequency and observation duration without requiring pointwise temporal alignment. In addition, PhyNet represents high-sampling-rate response signals as local response tokens embedded within time-frequency spectrograms, capturing localized spectral-temporal patterns associated with fault intensity and state transitions. To bridge heterogeneous token granularities, PhyNet introduces a learnable global response token that aggregates macroscopic information from response patches and serves as a compatible interface for physical variable interactions. Meanwhile, PhyNet develops a physical-to-response cross-attention module to inject constraints and causal cues from physical variable tokens into the global response token. Subsequently, the updated global token propagates these constraints and causal information to individual local response tokens via a self-attention mechanism. Without requiring strict alignment between physical variable sequences and response signals, PhyNet achieves physically constrained diagnosis, effectively mitigating information misalignment and the resulting noise while improving robustness across operating conditions. It provides a structured and novel route for fusing physical variables and response signals in deployment-oriented industrial diagnosis.

We evaluate PhyNet across four fluid-transport benchmarks. Our primary case study uses real-world cavitation datasets (Cavitation-Short/Long/Noise) provided by SAMSON AG, which pairs acoustic measurement responses with routinely recorded operational physical variables and covers multiple cavitation operating states under both noisy and noise-free conditions. Beyond cavitation phenomena, we further evaluate PhyNet on one additional fluid-transport benchmark with heterogeneous perception and operating-condition variations, namely VTPFR, an annular-flow image dataset capturing vertical gas-liquid two-phase flow regime dynamics and transitions. These benchmarks collectively cover representative monitoring tasks and diverse physical phenomena in industrial fluid transport systems, reflecting substantial variability arising from operational condition changes and noise-source interference. Across all scenarios, integrating physical variables as global constraints consistently improves diagnostic robustness without rigid alignment. More generally, PhyNet converts irregular and unaligned physical variables into actionable context for local response interpretation, providing a deployment-oriented solution for realistic fluid-transport monitoring. This enables reliable and robust fault diagnosis in fluid transport scenarios under realistic operational variability and offers potential for extension to other industrial settings.

\section*{Results}
\label{sec: 2_Results}
\begin{figure*}[htbp]
    \centering
    \includegraphics[width=\textwidth,height=140mm]{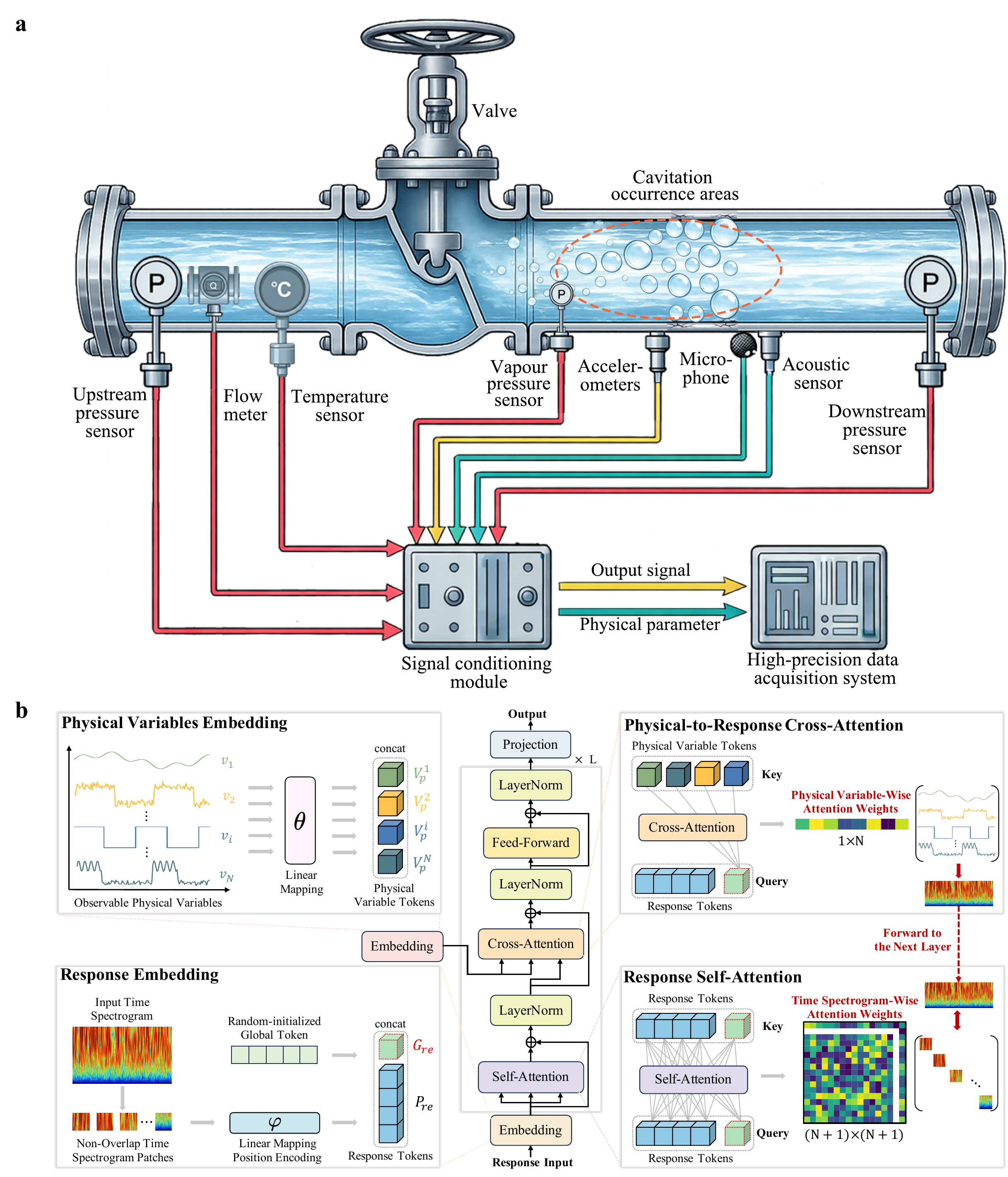}
    \caption{\textbf{Overview of PhyNet architecture}. \textbf{a}, Schematic of an industrial fluid transport system monitoring setup. High-frequency response measurements and routinely recorded physical variables are acquired by the corresponding sensors. These are synchronously processed through a signal conditioning module and a high-precision data acquisition system to form paired response signal-physical variable observations for diagnosis. \textbf{b}, Architecture of PhyNet. Physical Variables Embedding summarizes each variable sequence as a sequence-wise variate token. Response Embedding embeds non-overlapping patches as response tokens and generates a learnable global response token. Physical-to-Response Cross-Attention injects physical variate tokens into the global response token. Response Self-Attention propagates the updated global context to all response tokens. The attention blocks are stacked for $L$ layers.}
    \label{fig: OverallPhyNet}
\end{figure*}


\subsection*{Architecture of PhyNet}
\label{subsec: 2_1_architecture PhyNet}
Figure \ref{fig: OverallPhyNet}a illustrates heterogeneous monitoring in industrial fluid transport systems, where high-frequency responses and conventional physical variables are collected by different sensors. Although different modalities collectively characterise the system’s operational state, they are inherently heterogeneous in sampling frequency, observation duration and information granularity. To address this challenge, PhyNet adopts an asymmetric representation strategy, treating physical variables as carriers of global operating-condition constraints and regarding high-frequency response signals as local diagnostic evidence. As shown in Figure \ref{fig: OverallPhyNet}b, each physical-variable sequence is first encoded as a sequence-wise variate token, whereas the time-frequency representation of the response signal is partitioned into multiple local patches and encoded as response tokens. In addition, a learnable global response token is introduced to serve as a bridge for cross-modal interaction. Subsequently, physical variables are injected into the global response token via a physical-to-response cross-attention module, and the resulting global contextual information is then propagated back to the local response representation via a response self-attention mechanism, enabling physically informed diagnosis without requiring strict one-to-one alignment between modalities. This design treats physical variables as global context and mechanistic constraints that shape the interpretation of local time-frequency evidence under heterogeneous monitoring conditions. Consequently, the two sensing streams retain their original observational structures rather than being forced into a common temporal grid, which is important when their sampling rates and durations differ substantially.

\subsection*{Experiments and Performance}
\label{subsec: 2_2_Experiments}
We systematically evaluate PhyNet on four representative datasets, including three real-world cavitation datasets from SAMSON AG (Cavitation-Short, Cavitation-Long and Cavitation-Noise) and one additional public fluid transport dataset (VTPFR). These datasets encompass acoustic measurements, image observations, different noise levels and substantial operating-condition variation, providing a testing environment for assessing model generalization in heterogeneous industrial settings. In addition, the experimental comparisons incorporate three categories of representative baselines, comprising data-driven, physics-driven and hybrid data-physics methods. A uniform data-splitting strategy is applied to all datasets, with 80.00$\%$ of the samples used for training and the remaining 20.00$\%$ for testing. To prevent data leakage, sliding-window augmentation is performed only after the training and testing sets have been separated. Benchmark, experimental setting and metric details are provided in the ``Methods" section. 
\begin{table}[htbp]
\caption{Comparison of model performance across four datasets.}
\label{tab: OverallResults}
\centering
\fontsize{5}{6}\selectfont
\setlength{\tabcolsep}{0.3mm}
\begin{tabular}{llcccccccccccccccc}
\toprule
\multirow{2}{*}{\textbf{Types}} & \multirow{2}{*}{\textbf{Methods}} 
& \multicolumn{4}{c}{\textbf{Cavitation-Short}}                                                                    
& \multicolumn{4}{c}{\textbf{Cavitation-Long}}                                                                     
& \multicolumn{4}{c}{\textbf{Cavitation-Noise}}                                                                    
& \multicolumn{4}{c}{\textbf{VTPFR}}                                                                               
\\ 
\cmidrule{3-18} 
& 
& \multicolumn{1}{c}{Acc.} 
& \multicolumn{1}{c}{Pre.} 
& \multicolumn{1}{c}{Rec.} 
& \multicolumn{1}{c}{F1} 

& \multicolumn{1}{c}{Acc.} 
& \multicolumn{1}{c}{Pre.} 
& \multicolumn{1}{c}{Rec.} 
& \multicolumn{1}{c}{F1} 

& \multicolumn{1}{c}{Acc.} 
& \multicolumn{1}{c}{Pre.} 
& \multicolumn{1}{c}{Rec.} 
& \multicolumn{1}{c}{F1} 

& \multicolumn{1}{c}{Acc.} 
& \multicolumn{1}{c}{Pre.} 
& \multicolumn{1}{c}{Rec.} 
& \multicolumn{1}{c}{F1} 
\\
\midrule

\multirow{28}{*}{Data} 
& ResNet18
& 89.39 & 89.40 & 89.39 & 88.51
& 93.35 & 93.33 & 93.35 & 93.33
& 98.05 & 98.05 & 98.05 & 98.05
& 87.30 & 87.24 & 87.30 & 87.22 \\

& ResNet34
& 89.55 & 89.71 & 89.55 & 88.79
& 93.38 & 93.30 & 93.38 & 93.32
& 98.06 & 98.06 & 98.06 & 98.06
& 88.89 & 88.89 & 88.89 & 88.63 \\

& ResNet50
& 89.24 & 89.24 & 89.24 & 88.48
& \underline{93.44} & \underline{93.35} & \underline{93.44} & \underline{93.36}
& 97.84 & 97.84 & 97.84 & 97.84
& 84.66 & 84.56 & 84.66 & 84.36 \\

& DenseNet121
& \underline{90.11} & \underline{90.22} & \underline{90.11} & \underline{89.45}
& 93.30 & 93.17 & 93.30 & 93.19
& \underline{98.58} & \underline{98.58} & \underline{98.58} & \underline{98.58}
& 88.36 & 88.89 & 88.36 & 87.93 \\

& DenseNet161
& 89.94 & 89.99 & 89.94 & 89.23
& 93.40 & 93.23 & 93.40 & 93.27
& 97.76 & 97.76 & 97.76 & 97.76
& \underline{89.95} & \underline{89.93} & \underline{89.95} & \underline{89.82} \\

& ConvNeXt-T
& 87.60 & 87.20 & 87.60 & 87.04
& 92.93 & 92.86 & 92.93 & 92.87
& 97.96 & 97.96 & 97.96 & 97.96
& 72.49 & 72.07 & 72.49 & 71.59 \\

& ConvNeXt-S
& 87.45 & 87.71 & 87.45 & 86.11
& 92.67 & 92.70 & 92.67 & 92.66
& 98.48 & 98.48 & 98.48 & 98.48
& 71.96 & 71.83 & 71.96 & 71.11 \\

& ConvNeXt-B
& 88.20 & 88.07 & 88.20 & 87.37
& 92.62 & 92.61 & 92.62 & 92.59
& 97.80 & 97.80 & 97.80 & 97.80
& 71.96 & 71.91 & 71.96 & 71.34 \\

& ConvNeXt-L
& 87.99 & 88.14 & 87.99 & 87.37
& 93.12 & 93.00 & 93.12 & 93.04
& 97.48 & 97.48 & 97.48 & 97.48
& 74.07 & 73.53 & 74.07 & 73.61 \\
\cmidrule{2-18}

& ViT-T
& 89.59 & 90.04 & 89.59 & 88.66
& 92.21 & 92.04 & 92.21 & 92.11
& 98.33 & 98.33 & 98.33 & 98.33
& 73.54 & 73.43 & 73.54 & 73.27 \\

& ViT-S
& 88.90 & 88.92 & 88.90 & 87.70
& 92.53 & 92.50 & 92.53 & 92.49
& 98.27 & 98.27 & 98.27 & 98.27
& 74.07 & 74.09 & 74.07 & 74.06 \\

& ViT-B
& 89.26 & 89.80 & 89.26 & 88.24
& 92.23 & 91.99 & 92.23 & 92.09
& 97.80 & 97.80 & 97.80 & 97.80
& 75.13 & 75.43 & 75.13 & 74.10 \\

& Swin-T
& 89.37 & 89.47 & 89.37 & 89.07
& 93.41 & 93.24 & 93.41 & 93.26
& 98.64 & 98.64 & 98.64 & 98.64
& 79.37 & 80.00 & 79.37 & 79.33 \\

& Swin-S
& 89.31 & 89.04 & 89.31 & 88.90
& 92.61 & 92.45 & 92.61 & 92.46
& 98.40 & 98.40 & 98.40 & 98.40
& 76.19 & 76.30 & 76.19 & 76.18 \\

& Swin-B
& \underline{90.37} & 90.49 & \underline{90.37} & \underline{89.71}
& 91.46 & 91.10 & 91.46 & 91.20
& 97.76 & 97.76 & 97.76 & 97.76
& 78.84 & 79.33 & 78.84 & 78.94 \\

& PerViT-T
& 90.26 & \underline{90.54} & 90.26 & 89.53
& \underline{93.90} & \underline{93.78} & \underline{93.90} & \underline{93.82}
& \underline{98.68} & \underline{98.68} & \underline{98.68} & \underline{98.68}
& 82.01 & 82.02 & 82.01 & 81.28 \\

& PerViT-S
& 88.70 & 89.58 & 88.70 & 86.76
& 92.76 & 92.76 & 92.76 & 92.71
& 98.50 & 98.50 & 98.50 & 98.50
& \underline{85.19} & \underline{85.86} & \underline{85.19} & \underline{84.65} \\

& PerViT-M
& 88.78 & 88.71 & 88.68 & 87.79
& 93.20 & 92.97 & 93.20 & 93.04
& 97.86 & 97.86 & 97.86 & 97.86
& 84.13 & 84.11 & 84.13 & 83.68 \\

& CrossViT-T
& 89.46 & 89.91 & 89.46 & 88.39
& 92.25 & 91.96 & 92.25 & 92.06
& 98.51 & 98.51 & 98.51 & 98.51
& 80.42 & 80.58 & 80.42 & 80.41 \\

& CrossViT-S
& 89.87 & 89.75 & 89.87 & 89.04
& 92.48 & 92.35 & 92.48 & 92.41
& 98.63 & 98.63 & 98.63 & 98.63
& 81.48 & 81.76 & 81.48 & 81.59 \\

& CrossViT-B
& 89.76 & 89.86 & 89.76 & 88.89
& 92.31 & 92.31 & 92.31 & 92.26
& 98.31 & 98.31 & 98.31 & 98.31
& 83.07 & 83.23 & 83.07 & 82.71 \\
\cmidrule{2-18}

& UniFormer-S
& 88.81 & 89.45 & 88.81 & 88.01
& 93.22 & 93.13 & 93.22 & 93.16
& \underline{98.71} & \underline{98.71} & \underline{98.71} & \underline{98.71}
& 87.30 & 88.79 & 87.30 & 86.74 \\

& UniFormer-B
& 88.44 & 89.17 & 88.44 & 87.58
& 93.51 & 93.33 & 93.51 & 93.39
& 98.64 & 98.64 & 98.64 & 98.64
& 86.24 & 86.11 & 86.24 & 85.91 \\

& ConFormer
& 89.33 & \underline{90.14} & 89.33 & 88.33
& 93.03 & 92.85 & 93.03 & 92.91
& 98.68 & 98.68 & 98.68 & 98.68
& 87.30 & 88.14 & 87.30 & 86.72 \\

& CFFormer-S
& 89.24 & 89.47 & 89.24 & 88.53
& 93.53 & 93.48 & 93.53 & 93.45
& 98.62 & 98.62 & 98.62 & 98.62
& 88.36 & 89.24 & 88.36 & 87.87 \\

& CFFormer-B
& 88.87 & 88.86 & 88.87 & 88.29
& \underline{93.60} & \underline{93.53} & \underline{93.60} & \underline{93.54}
& 98.61 & 98.61 & 98.61 & 98.61
& 87.83 & 87.47 & 87.83 & 87.09 \\

& CFFormer-L
& \underline{89.68} & 89.67 & \underline{89.68} & \underline{88.97}
& 93.55 & 93.46 & 93.55 & 93.48
& 97.65 & 97.65 & 97.65 & 97.65
& \underline{89.42} & \underline{90.24} & \underline{89.42} & \underline{88.99} \\

& LeViT-S
& 88.83 & 88.61 & 88.83 & 87.98
& 93.05 & 92.95 & 93.05 & 92.97
& 98.63 & 98.63 & 98.63 & 98.63
& 84.66 & 84.33 & 84.66 & 82.80 \\

& LeViT-B
& 89.11 & 88.93 & 89.11 & 88.56
& 93.09 & 93.00 & 93.09 & 93.03
& 98.54 & 98.54 & 98.54 & 98.54
& 87.83 & 87.47 & 87.83 & 87.09 \\

\midrule

\multirow{4}{*}{Physics}
& SVM
& 88.57 & 82.77 & 80.76 & 81.07
& 90.96 & \underline{90.71} & \underline{90.96} & \underline{90.69}
& 99.18 & 99.18 & 99.18 & 99.18
& 71.96 & 70.44 & 74.97 & 71.28 \\

& DT
& 85.71 & 81.21 & 81.62 & 81.37
& 88.20 & 81.15 & 78.01 & 78.51
& 98.33 & 98.33 & 98.33 & 98.33
& 66.14 & 64.00 & 70.70 & 65.68 \\

& KNN
& 88.57 & 91.24 & \underline{82.10} & \underline{84.52}
& 91.93 & 90.56 & 89.60 & 89.80
& 98.47 & 98.47 & 98.47 & 98.47
& 70.37 & \underline{72.28} & 68.50 & 68.69 \\

& XGBoost
& \underline{90.00} & \underline{92.11} & 81.59 & 83.21
& \underline{93.17} & 89.09 & 87.70 & 88.30
& \underline{99.36} & \underline{99.36} & \underline{99.36} & \underline{99.36}
& \underline{73.02} & 71.75 & \underline{75.87} & \underline{72.39} \\

\midrule

\multirow{5}{*}{Hybrid}
& ViLT
& 90.63 & 91.58 & 90.63 & 89.39
& 92.10 & 92.09 & 92.10 & 92.01
& 99.19 & 99.20 & 99.19 & 99.19
& 91.53 & 92.25 & 91.53 & 91.43 \\

& Perceiver
& 91.21 & 90.99 & 91.21 & 90.69
& 93.98 & \underline{93.90} & 93.98 & \underline{93.92}
& 99.27 & 99.27 & 99.27 & 99.27
& 92.06 & 92.17 & 92.06 & 92.05 \\

& PerceiverIO
& 91.49 & 91.26 & 91.49 & 90.94
& \underline{94.03} & 93.80 & \underline{94.03} & 93.85
& \underline{99.40} & \underline{99.40} & \underline{99.40} & \underline{99.40}
& \underline{92.59} & \underline{93.75} & \underline{92.59} & \underline{92.71} \\

& MulT
& \underline{91.99} & \underline{92.51} & \underline{91.99} & \underline{91.32}
& 93.55 & 93.40 & 93.55 & 93.43
& 98.33 & 98.33 & 98.33 & 98.33
& 89.95 & 91.07 & 89.95 & 89.90 \\

\rowcolor{gray!15}
\cellcolor{white}
& \textbf{PhyNet}
& \textbf{93.72} & \textbf{93.58} & \textbf{93.72} & \textbf{93.57}
& \textbf{96.59} & \textbf{96.57} & \textbf{96.59} & \textbf{96.55}
& \textbf{100.0} & \textbf{100.0} & \textbf{100.0} & \textbf{100.0}
& \textbf{96.30} & \textbf{96.32} & \textbf{96.30} & \textbf{96.29} \\
\bottomrule
\end{tabular}
\vspace{1mm}
\begin{minipage}{\linewidth}
\fontsize{5}{6}\selectfont
Data, Physics and Hybrid denote data-driven methods, physics-driven methods and hybrid data-physics methods, respectively. Specifically, the data-driven methods include convolutional neural network (CNN)-based methods, Transformer-based methods and CNN-Transformer hybrid methods. Precision (Pre.), Recall (Rec.) and F1-score (F1) values are reported as weighted averages. For input preprocessing, DenseNet uses an input size of 256$\times$256, whereas all other models use 224$\times$224. For each dataset and metric, the best result is highlighted in bold, whereas the second-best result within the same method category is underlined. The results are averaged over three independent model-training runs using the same predefined training-test partition.
\end{minipage}
\end{table}

As shown in Table \ref{tab: OverallResults}, PhyNet achieves the best performance across all four datasets. It attains accuracies of 93.72$\%$, 96.59$\%$, 100.00$\%$ and 96.30$\%$ on Cavitation-Short, Cavitation-Long, Cavitation-Noise and VTPFR, respectively, while also ranking first in precision, recall and F1-score on all four datasets. This indicates that the advantage is not confined to a single evaluation measure or benchmark, but reflects a systematic improvement in overall discriminative performance across sensing modalities and tasks. 

Table \ref{tab: OverallResults} shows that the strongest data-driven baselines (Swin-B, PerViT-T, UniFormer-S and DenseNet161) achieve peak accuracies of 90.37$\%$, 93.90$\%$, 98.71$\%$ and 89.95$\%$ on Cavitation-Short, Cavitation-Long, Cavitation-Noise and VTPFR, respectively, which are 3.35$\%$, 2.81$\%$, 1.29$\%$ and 6.35$\%$ lower than PhyNet. For example, on Cavitation-Short, PhyNet improves precision, recall and F1-score over the strongest data-driven baseline (PerViT-T and Swin-B) by 3.04$\%$, 3.35$\%$ and 3.86$\%$, respectively. On VTPFR, PhyNet attains precision, recall and F1-score of 96.32$\%$, 96.30$\%$ and 96.29$\%$, representing improvements of 6.08$\%$, 6.35$\%$ and 6.47$\%$ over the strongest data-driven baseline (CFFormer-L and DenseNet161), respectively. These gains indicate that response-only models can learn local patterns and temporal dependencies but remain more susceptible to non-stationary effects under operating-condition variation, noise interference and modality heterogeneity. By contrast, PhyNet interprets local response evidence within a more stable physical context, leading to consistent improvements across all four metrics. 

Traditional machine learning methods based solely on physical variables reach best accuracies of 90.00$\%$, 93.17$\%$, 99.36$\%$ and 73.02$\%$ across the four datasets, which are consistently 0.64$\%$ to 23.28$\%$ lower than those of PhyNet. For example, on Cavitation-Long, PhyNet improves upon the strongest physics-driven baseline by 5.86$\%$, 5.63$\%$ and 5.86$\%$ in precision, recall and F1-score, respectively. On VTPFR, PhyNet outperforms the strongest physics-driven baseline by 24.04$\%$, 20.43$\%$ and 23.90$\%$ in precision, recall and F1-score, respectively. Thus, physical variables provide useful operating context and clearer interpretability but are insufficient alone for fine-grained localized state identification. PhyNet addresses this limitation by retaining discriminative response evidence while using physical variables to constrain its interpretation. 

Although hybrid data-physics methods (ViLT, Perceiver, PerceiverIO and MulT) also employ both original response measurements and physical variables, their best accuracies remain at 91.99$\%$, 94.03$\%$, 99.40$\%$ and 92.59$\%$ across four datasets, which are 1.73$\%$, 2.56$\%$, 0.60$\%$ and 3.71$\%$ lower than PhyNet, respectively. Meanwhile, the strongest hybrid baselines (MulT, Perceiver and PerceiverIO) reach F1-scores of 91.32$\%$, 93.92$\%$, 99.40$\%$ and 92.71$\%$ on Cavitation-Short, Cavitation-Long, Cavitation-Noise and VTPFR, all below the corresponding values of 93.57$\%$, 96.55$\%$, 100.00$\%$ and 96.29$\%$, respectively. These results indicate that the critical factor is not simply ``whether physical variables are introduced", but ``how they are incorporated". Existing hybrid models can process both modalities, but they often still rely implicitly or explicitly on aligned inputs or unified granularity. PhyNet instead uses a global response token and physical-to-response cross-attention to inject physical constraints at the representation level, reducing information mismatch and fusion noise under heterogeneous sampling.  

In particular, the improvements of PhyNet are more pronounced on Cavitation-Long, Cavitation-Noise and VTPFR, which are more susceptible to operational drift, prolonged observation durations or complex perceptual variation. These results show that physical variables not only serve as a supplementary information source under heterogeneous industrial observation conditions, but also as a manipulable global operating context that guides the interpretation of local response representations. The improvement therefore reflects a fusion strategy tailored to real-world sampling heterogeneity rather than the simple addition of another modality. 
\begin{figure*}[htbp]
    \centering
    \includegraphics[width=\textwidth,height=50mm]{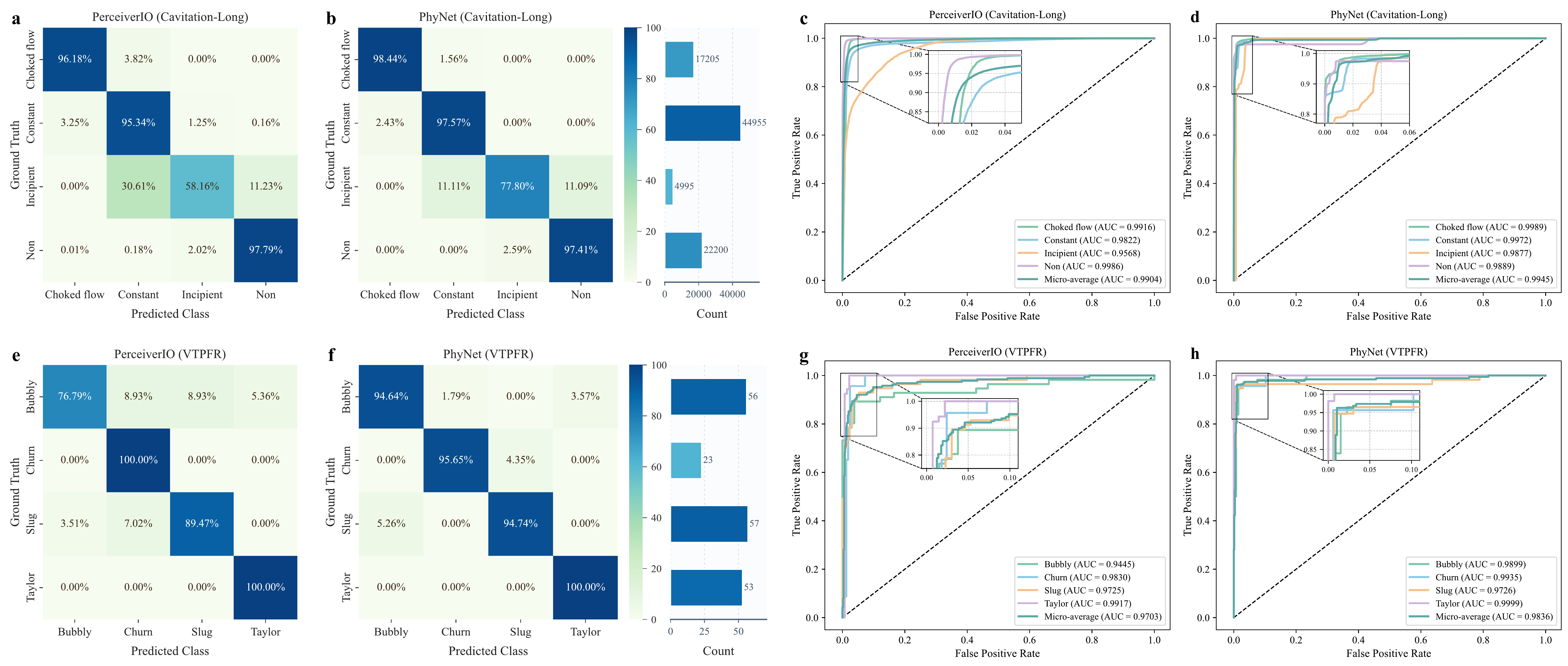}
    \caption{\textbf{Class-wise error patterns and ROC analysis on representative datasets}. \textbf{a, b}, Confusion matrices of PerceiverIO and PhyNet on Cavitation-Long, respectively. \textbf{c, d}, Receiver operating characteristic (ROC) curves of PerceiverIO and PhyNet on Cavitation-Long, respectively. \textbf{e, f}, Confusion matrices of PerceiverIO and PhyNet on VTPFR, respectively. \textbf{g, h}, ROC curves of PerceiverIO and PhyNet on VTPFR, respectively. The confusion matrices visualize class-wise prediction patterns and misclassification at the chosen decision threshold, whereas the ROC curves illustrate the relationship between the true positive rate and the false positive rate across thresholds. The area under each ROC curve corresponds to the AUC value, which provides a threshold-independent measure of discriminative performance.}
    \label{fig: ConfusionMatrixROC}
\end{figure*}

\begin{table}[htbp]
\caption{Comparison with domain-specific fault intensity diagnosis methods.}
\label{tab: Comparsion_cavitation_domain_specific}
\centering
\footnotesize
\setlength{\tabcolsep}{0.3mm}
\begin{tabular}{llcccccccccccc}
\toprule
\multicolumn{1}{c}{\multirow{2}{*}{Types}} & 
\multicolumn{1}{c}{\multirow{2}{*}{Methods}} & 
\multicolumn{4}{c}{Cavitation-Short} & 
\multicolumn{4}{c}{Cavitation-Long} & 
\multicolumn{4}{c}{Cavitation-Noise} \\ 
\cmidrule{3-14} 
\multicolumn{1}{c}{} & 
\multicolumn{1}{c}{} 
& Acc.    & Pre.    & Rec.    & F1     
& Acc.    & Pre.    & Rec.    & F1     
& Acc.    & Pre.    & Rec.    & F1     \\ 
\midrule
\multirow{4}{*}{Data}                      
& LiftingNet                                   
& 85.29   & 82.75   & 75.89   & 73.95  
& 88.43   & 86.31   & 87.97   & 87.05  
& 95.86   & 94.62   & 95.84   & 95.20  \\
                                           
& MIPLCNet                                     
& 86.57   & 83.38   & 77.71   & 75.00  
& 89.14   & 87.06   & 88.18   & 87.55  
& 96.57   & 95.78   & 96.61   & 96.18  \\
                                           
& ResNet-APReLU                                
& 86.86   & 84.04   & 77.89   & 75.56  
& 90.71   & 88.83   & 90.91   & 89.70  
& 97.29   & 96.77   & 97.49   & 97.12  \\
                                           
& LSTM-RDRN                                    
& \underline{87.71}   & \underline{85.42}   & \underline{78.50}   & \underline{76.72}  
& \underline{91.14}   & \underline{89.36}   & \underline{91.55}   & \underline{90.32}  
& \underline{98.71}   & \underline{98.40}   & \underline{98.53}   & \underline{98.46}  \\ 

\midrule
\multirow{5}{*}{Hybrid}                    
& HKG-ResNet34                                 
& 89.71   & 90.54   & 78.02   & 76.74  
& 92.44   & 88.55   & 92.21   & 90.16  
& 99.06   & 99.06   & 99.06   & 99.06  \\
                                           
& HKG-Swin-B                                   
& 89.57   & \underline{92.35}   & 77.88   & 76.66  
& 93.18   & 89.58   & 93.40   & 91.27  
& 99.63   & 99.63   & 99.63   & 99.62  \\
                                          
& DHK+CoNeXt-B                                 
& \underline{92.14}   & 90.71   & \underline{91.80}   & \underline{91.23}  
& \underline{94.35}   & \underline{91.03}   & 94.30   & \underline{92.52}  
& \underline{99.81}   & \underline{99.81}   & \underline{99.81}   & \underline{99.81}  \\

& DHK+Swin-B                                   
& 91.86   & 89.87   & 92.23   & 90.94  
& 94.19   & 90.88   & \underline{94.43}   & 92.45  
& 99.75   & 99.75   & 99.75   & 99.75  \\

\rowcolor{gray!15}
\cellcolor{white}
& \textbf{PhyNet}                                     
& \textbf{93.72}   & \textbf{93.58}   & \textbf{93.72}   & \textbf{93.57}  
& \textbf{96.59}   & \textbf{96.57}   & \textbf{96.59}   & \textbf{96.55}  
& \textbf{100.0}   & \textbf{100.0}   & \textbf{100.0}   & \textbf{100.0}  \\ 
\bottomrule
\end{tabular}
\vspace{1mm}
\begin{minipage}{\linewidth}
\fontsize{5}{6}\selectfont
Data and Hybrid denote data-driven methods and hybrid data-physics methods, respectively. Precision (Pre.), Recall (Rec.) and F1-score (F1) values are reported as weighted averages. For input preprocessing, LiftingNet \cite{pan2017liftingnet}, MIPLCNet \cite{pan2019novel}, ResNet-APReLU \cite{zhao2020deep}, LSTM-RDRN \cite{mohammad2023one} and HKG-ResNet34 \cite{yusha2024hierarchical} use an input size of 256$\times$256. HKG-Swin-B \cite{yusha2024hierarchical} and DHK+Swin-B \cite{yusha2026deep} use an input size of 384$\times$384, whereas all other models use 224$\times$224. For each dataset and metric, the best result is highlighted in bold, whereas the second-best result within the same method category is underlined. The results are averaged over three independent model-training runs using the same predefined training-test partition.
\end{minipage}
\end{table}

Table \ref{tab: Comparsion_cavitation_domain_specific} compares PhyNet with representative domain-specific and knowledge-guided methods. PhyNet achieves the best results across all metrics on the three cavitation datasets, with the clearest gains on Cavitation-Long: 2.24$\%$, 5.54$\%$, 2.16$\%$ and 4.03$\%$ in accuracy, precision, recall and F1-score, respectively. On Cavitation-Noise, it reaches 100.00$\%$ across all metrics. These results show that physical-variable constraints complement response-pattern learning and class-hierarchy regularization under long-duration and varying operating conditions. 

\subsection*{Mechanistic and robustness analysis}
\label{subsec: 2_3_Mechanistic and robustness analysis}

\subsubsection*{Class-wise errors and ROC curves}
\label{subsubsec: 2_3_1_Class-wise errors and ROC curves}
To further determine whether the performance improvement of PhyNet arises from better class-level discrimination, we compare its confusion matrices and ROC curves with those of the strongest hybrid baseline (PerceiverIO) on two representative datasets (Cavitation-Long and VTPFR) with longer observation durations, more pronounced operational variations or more complex perceptual structures, as shown in Figure \ref{fig: ConfusionMatrixROC}. Corresponding results for the other datasets are provided in the Supplementary Information.

The confusion matrices reveal that the performance gains of PhyNet are concentrated on the most ambiguous and error-prone classes. On Cavitation-Long, both models achieve high accuracy in classifying choked flow cavitation, constant cavitation and non-cavitation, as shown in Figures \ref{fig: ConfusionMatrixROC}a and \ref{fig: ConfusionMatrixROC}b. However, PerceiverIO exhibits substantial confusion for incipient cavitation with only 58.16$\%$ of samples correctly assigned to the diagonal and 30.61$\%$ misclassified as constant cavitation. By contrast, PhyNet increases the correct recognition rate of incipient cavitation to 77.89$\%$ and reduces the corresponding confusion with constant cavitation to 11.11$\%$, indicating a markedly improved separation of early and adjacent cavitation states under prolonged observation and operating-condition variation. On VTPFR, the largest improvement is observed for the bubbly regime, whose diagonal entry increases from 76.79$\%$ under PerceiverIO to 94.64$\%$ under PhyNet. A similar improvement is observed for the slug regime, which rises from 89.47$\%$ to 94.74$\%$, whereas the Taylor bubble regime maintains perfect recognition under both models. This illustrates that the primary advantage of PhyNet lies in reducing confusion among visually or physically adjacent states to mitigate interference from these states under heterogeneous operating conditions. 

The ROC curves further support this conclusion by providing a threshold-independent view of discriminative performance. For Cavitation-Long, the AUC for incipient cavitation increases from 0.9568 with PerceiverIO to 0.9877 with PhyNet, whereas the other classes remain close to the theoretical ceiling under both methods. In addition, the micro-average AUC improves from 0.9904 to 0.9945. For VTPFR, PhyNet exhibits superior ROC performance, with the largest improvement observed for the bubbly regime, whose AUC increases from 0.9445 to 0.9899 and the micro-average AUC rises from 0.9703 to 0.9836. These results indicate that the advantage of PhyNet is not tied to a specific classification threshold, but reflects more robust class separability over a broader range of decision thresholds.

\subsubsection*{Condition-aware attention}
\label{subsubsec: 2_3_2_Condition-aware attention}
To elucidate how physical variables modulate response representations, we examine the physical-variable attention patterns learned by the physical-to-response cross-attention module on Cavitation-Long, as shown in Figure \ref{fig: CrossAttention}. Representative high-confidence and low-margin boundary samples from choked-flow cavitation, constant cavitation, incipient cavitation and non-cavitation show that PhyNet consistently assigns the highest weights to vapour pressure $p_v$, with additional contributions from volumetric flow rate $Q$ and pressure recovery factor ${X_f}$, as shown in Figures \ref{fig: CrossAttention}a and \ref{fig: CrossAttention}b. This core pattern remains clearly visible even in boundary samples, indicating that the model relies on a set of physically meaningful operating variables when distinguishing between adjacent and difficult states. The class-averaged statistics in Figure \ref{fig: CrossAttention}c further confirm that vapour pressure $p_v$ dominates across all flow states, whereas the relative contributions of volumetric flow rate $Q$ and pressure recovery factor ${X_f}$ vary with flow conditions. This observation is highly consistent with the physical mechanisms of cavitation. Cavitation is initiated when the local static pressure falls to or below the vapour pressure $p_v$, making the vapour pressure $p_v$ a direct thermodynamic threshold for bubble inception. The volumetric flow rate $Q$ strongly affects local velocity and pressure drop through the constriction and therefore modulates the likelihood and intensity of cavity formation. The pressure recovery factor ${X_f}$ characterizes the relationship between upstream pressure $p_u$, downstream pressure $p_d$ and vapour pressure $p_v$, which is widely used to describe the operating pressure condition associated with cavitation onset and development. Therefore, the dominant contribution of vapour pressure $p_v$ and the secondary contribution of volumetric flow rate $Q$ and pressure recovery factor ${X_f}$ are not only findings supported by the data, but also consistent with established hydrodynamic cavitation theory \cite{kumar1999modeling,ranade2022modeling} (A detailed definition of cavitation is provided in the Supplementary Information). Head-wise visualizations further show that these averaged patterns arise from structured head specialization rather than from uniform weighting across heads (see Supplementary Information).
\begin{figure*}[htbp]
    \centering
    \includegraphics[width=\textwidth,height=105mm]{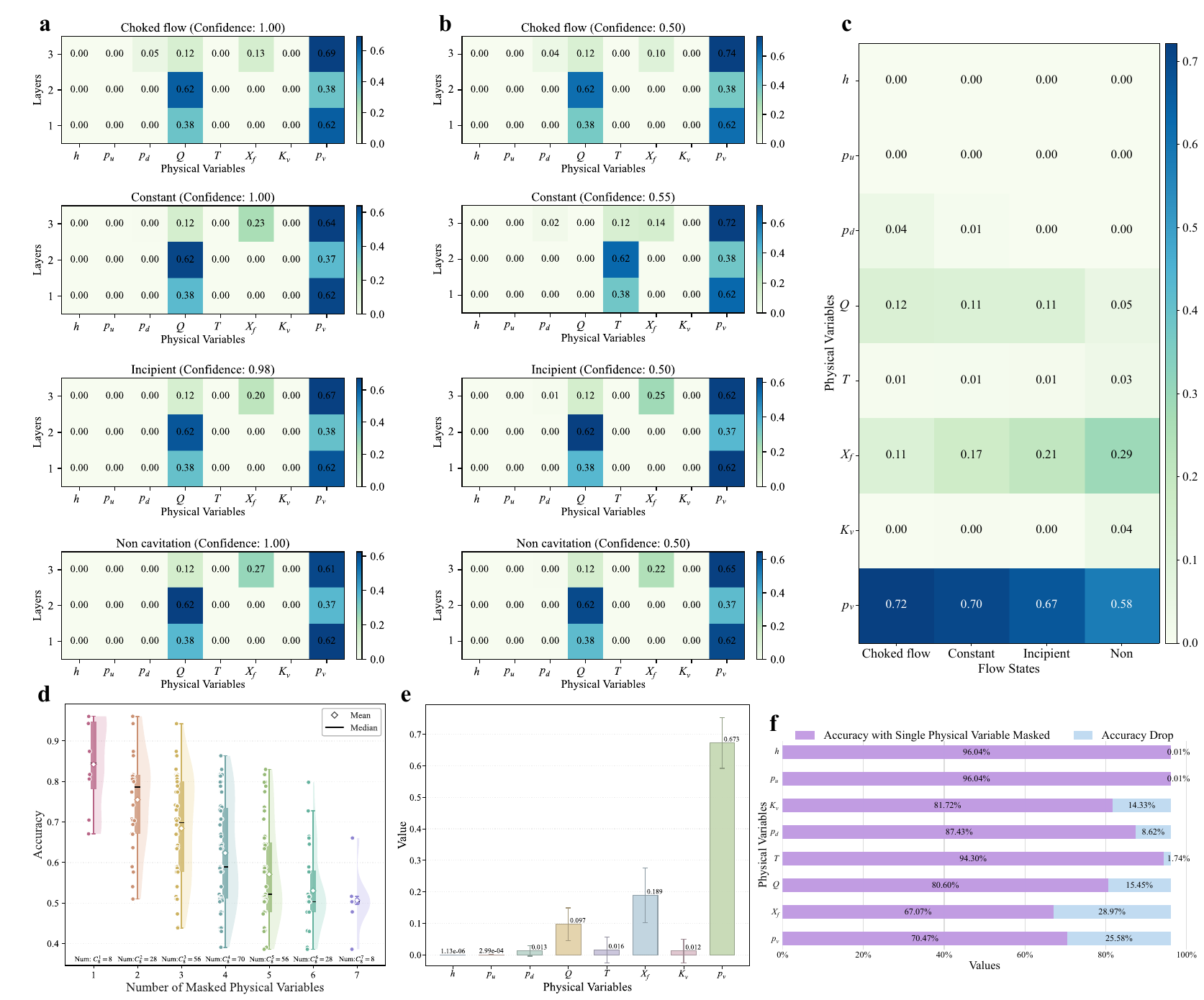}
    \caption{\textbf{Physical-response cross-attention and its association with diagnostic performance}. \textbf{a}, \textbf{b}, Layer-wise mean cross-attention weights from the global response token to physical-variable tokens for representative high-confidence prototype samples (a) and low-margin boundary samples (b) from the four flow states. High-confidence prototype samples denote correctly classified examples with high prediction confidence, whereas low-margin boundary samples denote correctly classified examples with small margins between the top two predicted classes. \textbf{c}, Class-averaged physical-variable attention across the four flow states. \textbf{d}, Accuracy under random masking of physical-variable combinations. Random masking of different numbers of physical-variable combinations is equivalent to sampling physical-variable subsets of different sizes. \textbf{e}, Physical-variable importance across the whole test set of Cavitation-Long. \textbf{f}, Accuracy under single-variable masking. The accuracy drop is defined relative to the 96.59$\%$ accuracy of the PhyNet (full model) on Cavitation-Long. The reported values are averaged over three independent model-training runs ($n=3$), and the error bars denote one standard deviation.}
    \label{fig: CrossAttention}
\end{figure*}

To determine whether these attention patterns are functionally relevant, we further relate them to diagnostic performance under physical-variable masking. As shown in Figure \ref{fig: CrossAttention}d, the accuracy declines progressively as increasing numbers of physical variables are randomly masked, with the median falling from 0.95 when one or two physical variables are masked to approximately 0.50 when seven variables are masked. This indicates that PhyNet's diagnosis depends on the overall physical context rather than any single variable. The mean importance of physical variables further ranks vapour pressure $p_v$ as the dominant variable (0.673), followed by the pressure recovery factor ${X_f}$ (0.189) and volumetric flow rate $Q$ (0.107), whereas the remaining variables contribute only marginally, as shown in Figure \ref{fig: CrossAttention}e. In addition, Figure \ref{fig: CrossAttention}f shows that single-variable masking causes the largest performance drops when the pressure recovery factor ${X_f}$, vapour pressure $p_v$ or volumetric flow rate $Q$ is masked, reducing accuracy from 96.59$\%$ to 67.07$\%$, 70.47$\%$ and 80.60$\%$, respectively. By contrast, masking valve stroke $h$, upstream pressure $p_u$ and fluid temperature $T$, results in only modest degradation. These results further demonstrate that the physical-to-response cross-attention module captures functionally relevant and physically meaningful operating constraints rather than relying on a single dominant variable, and that its learned importance pattern is consistent with cavitation physics.  
\begin{figure*}[htbp]
    \centering
    \includegraphics[width=\textwidth,height=155mm]{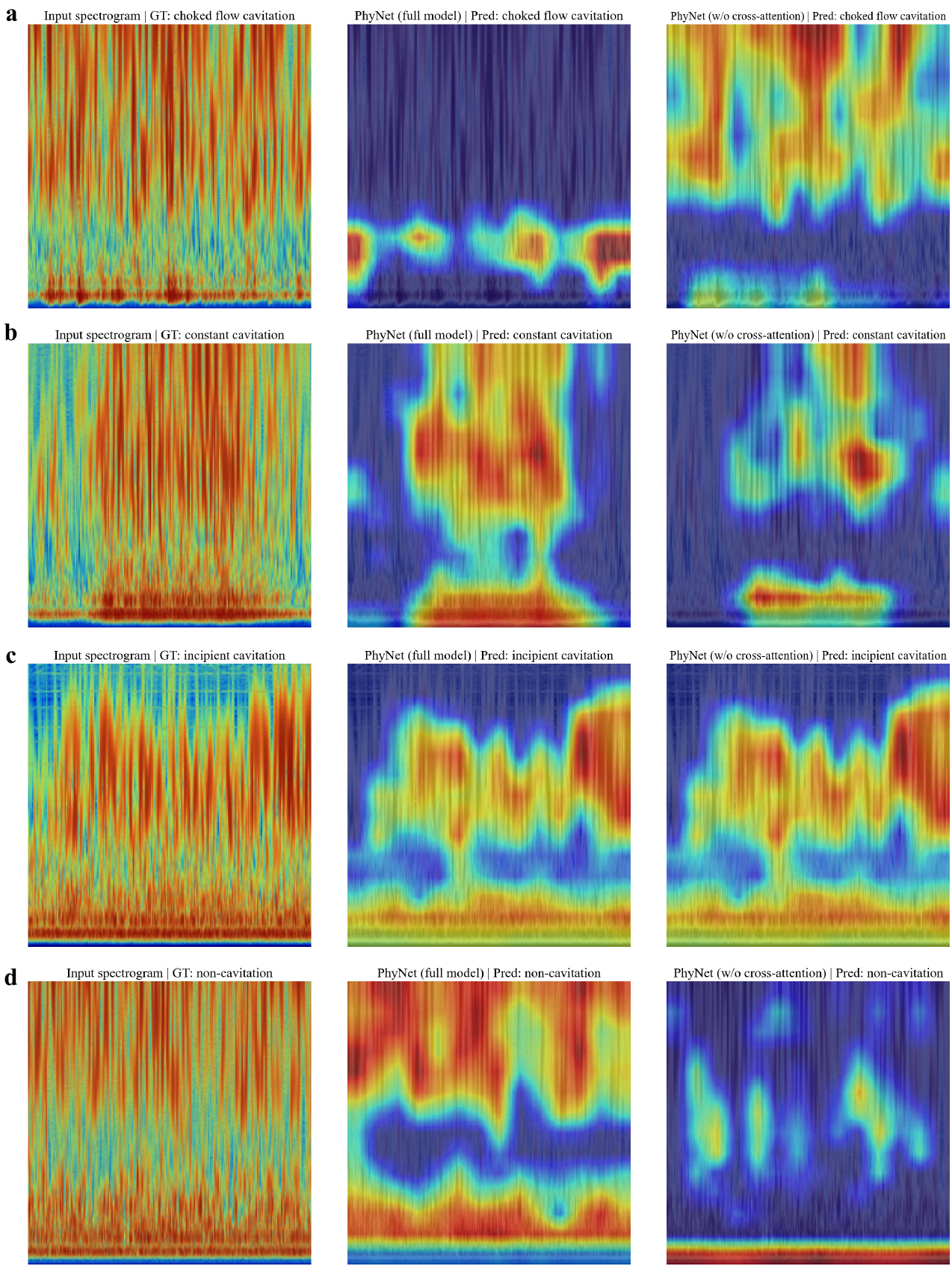}
    \caption{\textbf{Local response evidence under physical variables}. \textbf{a-d}, Representative spectrograms from choked-flow cavitation (a), constant cavitation (b), incipient cavitation (c) and non-cavitation (d). For each row, the left panel shows the input spectrogram, while the middle and right panels show the gradient-based localization maps produced by the complete PhyNet and PhyNet without physical-to-response cross-attention, respectively. All localization maps are displayed using the same colour scale. Technical details of the localization procedure are provided in the Supplementary Methods.}
    \label{fig: SelfAttention}
\end{figure*}

Furthermore, the impact of physical variable constraints is also evident at the local response level. We present gradient-based localization maps on spectrograms of Cavitation-Long for the full model and PhyNet without physical-to-response cross-attention, as shown in Figure \ref{fig: SelfAttention}. In the full model, the local response patterns maintain a broader and more structured correspondence with the time-frequency evidence associated with the target state. In contrast, PhyNet without cross-attention causes the response patterns to shift or fragment. Moreover, this change is often accompanied by an erroneous bias towards the non-cavitation class for cavitation samples. This suggests that physical variables not only provide auxiliary decision cues, but also actively reshape both the distribution and interpretation of local response evidence.

\subsubsection*{Asymmetric fusion analysis}
\label{subsubsec: 2_3_3_Asymmetric fusion analysis}
To determine whether the performance improvement of PhyNet stems from the proposed asymmetric fusion strategy, we design a series of ablation experiments on the Cavitation-Long dataset by systematically removing or simplifying key components of PhyNet, as shown in Table \ref{tab: AsymmetricFusion}. The full PhyNet model achieves an accuracy of 96.59$\%$ and an F1-score of 96.55$\%$, whereas all simplified variants exhibit substantial performance degradation. This indicates that the superior performance of PhyNet depends on the coordinated interplay among physical variables, the global response token and the structured cross-modal conditioning mechanism.
\begin{table}[htbp]
\caption{Asymmetric fusion ablation on Cavitation-Long.}
\label{tab: AsymmetricFusion}
\centering
\setlength{\tabcolsep}{6mm}
\begin{tabular}{ccccc}
\toprule
Variants                       & Acc. & Pre. & Rec. & F1 \\ 
\midrule
w/o physical variables         &79.01      &78.48      &79.01      &78.47    \\
w/o global response token      &85.71      &86.15      &85.71      &85.29    \\
w/o cross-attention            &87.88      &88.21      &87.88      &88.03    \\
w/o self-attention propagation &87.95      &87.35      &87.95      &87.60    \\
naive fusion                   &83.76      &84.21      &83.76      &82.85    \\
\rowcolor{gray!15}
PhyNet                         &\textbf{96.59}      &\textbf{96.57}      &\textbf{96.59}      &\textbf{96.55}    \\ 
\bottomrule
\end{tabular}
\vspace{1mm}
\begin{minipage}{\linewidth}
\footnotesize
Detailed definitions and implementation settings of all ablation variants are provided in the Supplementary Methods. Acc., Pre., Rec. and F1 denote accuracy, precision, recall and F1-score, respectively. All reported values are averaged over three experimental runs, with precision, recall and F1-score calculated as weighted averages within each run.
\end{minipage}
\end{table}

The without physical variables variant reduces accuracy and F1-score to 79.01$\%$ and 78.47$\%$, corresponding to declines of 17.58$\%$ and 18.08$\%$ compared to PhyNet. This result suggests that physical variables are not a marginal supplement to the response branch, but rather a critical source of global operating context for the diagnostic process. A similarly strong degradation is observed for naive fusion, which attains only 83.76$\%$ accuracy and 82.85$\%$ F1-score, corresponding to declines of 12.83$\%$ and 13.70$\%$, respectively. Because this variant still utilises both modalities while removing the encoder-level asymmetric conditioning in PhyNet, this result indicates that the critical factor is not simply whether physical variables are introduced, but how they are incorporated. In particular, late fusion without token-level conditional processing is insufficient to recover the benefits of physically informed response interpretation.

The remaining ablation experiments further elucidate the role of the internal interaction mechanism. The variant without the learnable global response token reduces the accuracy to 85.71$\%$ and the F1-score to 85.29$\%$, indicating that direct coupling between physical variable tokens and local response tokens is less effective than the semantically compatible global-interface interaction. The physical-response cross-attention module replaced with a simple injection module also results in an accuracy reduction to 87.88$\%$ and an F1-score decline to 88.03$\%$, indicating that explicit cross-attention is necessary for effective physical conditioning. Conversely, when the propagation process from global constraints back to local response tokens in the response self-attention mechanism is removed, the accuracy and F1-score of the model decrease to 87.95$\%$ and 87.60$\%$, respectively. The comparable degradation of the without cross-attention and without self-attention propagation variants suggests that simply injecting physical constraints into the global token is insufficient. These constraints must also be redistributed to local time-frequency evidence to achieve robust diagnostics. In summary, these results collectively support PhyNet's central asymmetric fusion strategy, whereby physical variables are introduced into the model as global constraints via a dedicated semantic bridge and subsequently propagated back to local response representations. Compared with fusion applied only at the classifier stage or with the injection of physical information through simplified global updates, the strategy of PhyNet enables the diagnostic value of physical variables to be exploited more effectively.

\subsubsection*{Robustness analysis}
\label{subsubsec: 2_3_4_Robustness analysis}
To further assess the robustness of PhyNet under variations in signal preprocessing, model configuration and acquisition conditions, we conduct a series of sensitivity experiments, as shown in Figure \ref{fig: RobustnessAnalysis}. First, we report the influence of time-frequency preprocessing on Cavitation-Short, as shown in Figures \ref{fig: RobustnessAnalysis}a-\ref{fig: RobustnessAnalysis}c. In the window duration experiments, a window duration of \SI{0.06}{s} obtains the highest accuracy of 93.72$\%$, whereas substantially shorter or longer windows reduce performance to 86.65$\%$ at \SI{0.01}{s} and 90.15$\%$ at \SI{0.3}{s}, respectively (see Figure \ref{fig: RobustnessAnalysis}a). A similar trend is observed for the window overlap ratio (see Figure \ref{fig: RobustnessAnalysis}b), where the default value of 0.25 achieves the best accuracy of 93.72$\%$ compared with 91.85$\%$ at 0.50 and 82.46$\%$ at 0.75. In addition, the results of varying the number of continuous wavelet transform (CWT) scales indicate that 512 scales provide the most effective time-frequency representation with an accuracy of 93.72$\%$, whilst 128, 256 and 1024 scales achieve 89.89$\%$, 91.08$\%$ and 92.38$\%$, respectively (see Figure \ref{fig: RobustnessAnalysis}c). In summary, these results indicate that the diagnostic performance of PhyNet is robust within a reasonable range of preprocessing options, whilst overly coarse or overly redundant time-frequency representations can degrade its performance.
\begin{figure*}[htbp]
    \centering
    \includegraphics[width=\textwidth,height=105mm]{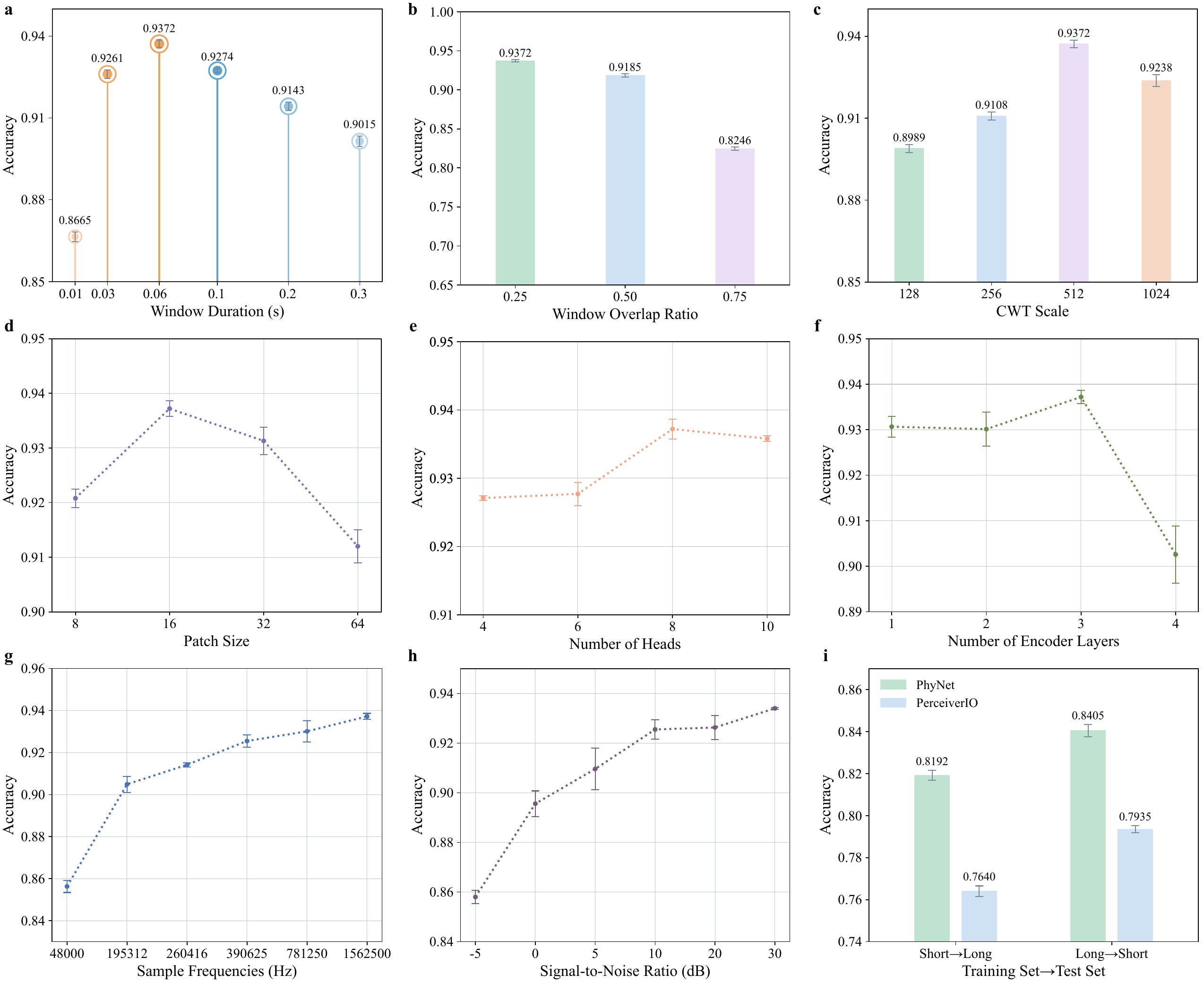}
    \caption{\textbf{Robustness analysis of PhyNet under various conditions}. \textbf{a-c}, Sensitivity of PhyNet to time-frequency preprocessing on Cavitation-Short, including window duration (a), window overlap ratio (b) and number of continuous wavelet transform (CWT) scales (c). \textbf{d-f}, Sensitivity of PhyNet to architectural hyperparameters on Cavitation-Short, including patch size (d), number of attention heads (e) and number of encoder layers (f). \textbf{g}, Performance of PhyNet across different sampling frequencies on Cavitation-Short. \textbf{h}, Accuracy of PhyNet under different signal-to-noise ratios on Cavitation-Short. \textbf{i}, Performance comparison between PhyNet and the strongest hybrid baseline (PerceiverIO) during cross-duration transfer between Cavitation-Short and Cavitation-Long (Training set$\to$Test set: Short$\to$Long and Long$\to$Short). Panels \textbf{a-h} show the performance of PhyNet under isolated perturbations of a single factor, while the remaining settings are fixed at their default values. The reported values are averaged over three independent model-training runs ($n=3$), and the error bars denote one standard deviation. In addition, the results for the remaining metrics are provided in the Supplementary Information.}
    \label{fig: RobustnessAnalysis}
\end{figure*}

Second, we evaluate the sensitivity of PhyNet to architectural hyperparameters on Cavitation-Short, as shown in Figures \ref{fig: RobustnessAnalysis}d-\ref{fig: RobustnessAnalysis}f. In terms of patch size, PhyNet achieves the best performance with a patch size of 16, outperforming patch sizes of 8, 32 and 64 by 1.64$\%$, 0.59$\%$ and 2.52$\%$, respectively (see Figure \ref{fig: RobustnessAnalysis}d). Similarly, PhyNet demonstrates relative stability across the optimisation range for the number of attention heads, as depicted in Figure \ref{fig: RobustnessAnalysis}e. Specifically, PhyNet attains the highest accuracy with 8 attention heads, whereas the accuracy with 4, 6 and 10 attention heads reduces by 1.01$\%$, 0.95$\%$ and 0.14$\%$, respectively. Furthermore, the encoder depth of PhyNet exhibits a comparable pattern, as illustrated in Figure \ref{fig: RobustnessAnalysis}f. PhyNet with 3 encoder layers obtains the best performance, remains broadly comparable with 1 or 2 encoder layers and declines markedly when the encoder depth is increased to 4 layers. These results illustrate that the proposed PhyNet does not rely on excessive architectural complexity and its moderate configuration is sufficient to effectively capture the interaction between global physical constraints and local response evidence.

Next, the ability to recognize signals captured by low-cost sensors is important and challenging in practical applications. Therefore, we evaluate PhyNet under the original sampling frequency (\SI{1562500}{Hz}), one-half, one-quarter, one-sixth, one-eighth of the original sampling frequency (\SI{781250}{Hz}, \SI{390625}{Hz}, \SI{260416}{Hz}, \SI{195312}{Hz}) and the maximum frequency typically supported by mobile phones ($\SI{48000}{Hz} \approx \frac{1562500}{32}\,\si{Hz}$), as shown in Figure \ref{fig: RobustnessAnalysis}g. As the sampling frequency increases from \SI{48000}{Hz} to \SI{1562500}{Hz}, the performance rises steadily from 85.63$\%$ to 93.72$\%$, indicating that finer resolution provides more informative diagnostic evidence and PhyNet demonstrates substantial robustness when data quality fluctuates. Moreover, we assess the robustness of PhyNet under additive noise on Cavitation-Short, as shown in Figure \ref{fig: RobustnessAnalysis}h. Specifically, PhyNet exhibits a consistent improvement as the signal-to-noise ratio increases, with its accuracy rising from 85.80$\%$ at \SI{-5}{dB} to 89.56$\%$ at \SI{0}{dB}, 90.96$\%$ at \SI{5}{dB}, 92.55$\%$ at \SI{10}{dB}, 92.63$\%$ at \SI{20}{dB} and 93.40$\%$ at \SI{30}{dB}. This trend indicates that PhyNet remains stable under moderate noise and only deteriorates substantially under very severe signal corruption.

Finally, we evaluate the cross-duration generalization capability of PhyNet between Cavitation-Short and Cavitation-Long, as shown in Figure \ref{fig: RobustnessAnalysis}i. When models are trained on short-duration observations and tested on long-duration observations, PhyNet achieves an accuracy of 81.92$\%$, representing a 5.52$\%$ improvement over PerceiverIO. Conversely, when models are trained on Cavitation-Long and tested on Cavitation-Short, PhyNet and PerceiverIO obtain accuracies of 84.05$\%$ and 79.35$\%$, respectively. This demonstrates that the proposed physical variable conditioning and asymmetric fusion strategies not only enhance model robustness to preprocessing and acquisition perturbations, but also improve its adaptability to variations in observation duration. In addition, PhyNet remains stable across a broad range of practical variations, providing a more deployment-oriented diagnosis framework for heterogeneous industrial environments.

\section*{Discussion}
\label{sec: 3_Discussion}
The results indicate that the advantage of PhyNet arises not simply from adding another modality, but from changing how heterogeneous modalities interact. Across four industrial fluid-transport benchmarks spanning acoustic and visual sensing, noisy and noise-free conditions, and substantial operating variation, PhyNet achieves the best performance on all evaluation metrics. The gains are particularly pronounced on Cavitation-Long, Cavitation-Noise and VTPFR, which are more affected by operational drift, prolonged observation or perceptual variation. These findings suggest that physical variables can provide a stable context for interpreting local response evidence when treated as global operating constraints rather than conventional tabular inputs. This distinction is important because the observed improvements are not confined to a single dataset or sensing modality, but recur across several forms of acquisition heterogeneity and operating-condition variation.

Several observations explain this behaviour. First, the largest improvements occur in ambiguous and error-prone states. On Cavitation-Long, the correct recognition rate for incipient cavitation increases from 58.16$\%$ with PerceiverIO to 77.89$\%$ with PhyNet, while the AUC rises from 0.9568 to 0.9877. On VTPFR, the correct rate for the bubbly regime increases from 76.79$\%$ to 94.64$\%$, accompanied by a substantial AUC improvement. Physical conditioning therefore contributes not only to average performance, but also to discrimination where local response evidence is weakest, most non-stationary or most easily confused with neighbouring states. In this sense, PhyNet reduces reliance on condition-specific response patterns and supports more physically grounded class separation.

Second, attention and masking analyses indicate that PhyNet learns physically meaningful constraints rather than arbitrary multimodal correlations. Physical-to-response cross-attention assigns the highest weights to vapour pressure $p_v$, with volumetric flow rate $Q$ and pressure recovery factor ${X_f}$ as the main secondary contributors. This pattern persists across high-confidence prototypes, low-margin boundary samples, class-level averages and specialized attention heads. Its functional relevance is supported by masking experiments: removing $p_v$, $Q$ or ${X_f}$ produces the largest performance degradation, whereas masking other variables has smaller effects. The agreement between learned importance and established cavitation mechanisms supports both the diagnostic value and mechanistic plausibility of the captured operating constraints. It also suggests that the model does not distribute attention uniformly across available variables, but selectively emphasizes variables that govern cavitation inception and development.

Third, physical variables influence not only global decisions but also the distribution of evidence within response representations. Compared with the model without physical-to-response cross-attention, full PhyNet produces broader and more structured localization patterns on representative spectrograms. Removing cross-attention fragments or shifts these patterns and often biases cavitation samples towards non-cavitation predictions. The ablation results reinforce this interpretation: removing physical variables, the global response token, explicit cross-attention or subsequent self-attention propagation all substantially reduces performance, while naive fusion remains inferior to the full model. Effective industrial multimodal diagnosis therefore depends not only on what information is incorporated, but also on whether it enters through a semantically compatible and structurally appropriate interaction pathway.

The robustness analysis further highlights the practical relevance of this design. PhyNet remains stable across reasonable choices of window duration, overlap ratio, CWT scales and architectural hyperparameters, and achieves its best results with moderate rather than excessive model complexity. Performance also degrades gradually under reduced sampling frequency and additive noise. Moreover, PhyNet retains a clear advantage over the strongest hybrid baseline during cross-duration transfer between Cavitation-Short and Cavitation-Long. These results indicate resilience to variations in sensing quality, observation duration and preprocessing, all of which commonly occur across industrial systems and deployment sites. Such stability is particularly relevant when high-end sensors, long observation windows or identical acquisition settings cannot be guaranteed in practice.

More broadly, PhyNet provides a route for multimodal learning under industrial sampling heterogeneity. Physical variables are often sparse, asynchronous or low-frequency, whereas response measurements are dense, high-frequency and locally informative. Existing multimodal architectures frequently assume aligned sequences or unified granularities, which are difficult to obtain in practice. PhyNet instead embeds physical variables as sequence-wise variate tokens and couples them with local response evidence through a global response token, incorporating mechanistic operating context without enforcing one-to-one correspondence. Although developed for fluid-transport diagnosis, this principle may extend to other industrial systems that jointly monitor slowly varying operating conditions and rapidly sampled responses. In such settings, global operating variables can constrain the interpretation of local vibration, acoustic, visual or electrical evidence without requiring artificial temporal synchronization.

Although the present results show that PhyNet achieves strong performance and mechanistic consistency, there are several limitations that warrant further discussion. The experiments and mechanistic analyses focus primarily on fluid-transport systems, and validation in other industrial domains is needed to establish broader generalizability. The current framework also formulates diagnosis as supervised classification and does not address open-set recognition, fault-progression modelling or early warning. Although attention, variable masking and response localization provide complementary evidence for the role of physical constraints, more granular intervention studies and uncertainty analyses are required to characterize causal effects and failure boundaries. Future work should therefore examine missing or corrupted physical variables, uncertainty-aware modelling and more diverse deployment environments. Overall, the findings support using physical variables as actionable global context rather than auxiliary side inputs or late-stage features, providing a deployment-oriented approach to robust diagnosis in heterogeneous industrial fluid-transport systems.

\section*{Methods}
\label{sec: 4_Methods}
\subsection*{Data sources}
\label{subsec: 4_1_Data sources}
All data used in this research are openly available to the public (See ``Data availability" section). The cavitation dataset is provided by SAMSON AG, comprising \textbf{Cavitation-Short}, \textbf{Cavitation-Long} and \textbf{Cavitation-Noise} (real noise). The cavitation conditions include incipient cavitation, constant cavitation and choked-flow cavitation, while the non-cavitation conditions comprise turbulent flow and no-flow state. All acoustic signals are sampled at \SI{1562.5}{kHz}. \textbf{Cavitation-Short} contains 356 samples, each with a duration of \SI{3}{s}. \textbf{Cavitation-Long} and \textbf{Cavitation-Noise} contain 806 and 160 samples, respectively, with each signal lasting \SI{25}{s}. In addition, the recorded physical variables include upstream pressure $p_u$ (\si{\bar\,(g)}), downstream pressure $p_d$ (\si{\bar\,(g)}), vapour pressure $p_v$ (\si{\bar\,(g)}), volumetric flow rate $Q$ (\si{\cubic\metre\per\hour}), pressure recovery factor ${X_f} = {{({p_u} - {p_d})} \mathord{\left/{\vphantom {{({p_u} - {p_d})} {({p_u} - {p_v})}}} \right.\kern-\nulldelimiterspace} {({p_u} - {p_v})}}$, fluid temperature $T$ (\si{\degreeCelsius}), valve flow coefficient $K_v$ (\si{\cubic\metre\per\hour}) and valve stroke $h$ (\si{\milli\metre}). 

To verify the scalability and generalization capability of the proposed PhyNet, extensive evaluations are further conducted across other industrial fluid transport systems. The vertical two-phase flow regimes in an annular image dataset (VTPFR) \cite{manikonda2025vertical} is collected in the 140-ft tower laboratory at the Petroleum Engineering Building of Texas A$\&$M University for flow regime recognition. The dataset contains 947 high-resolution gas-liquid two-phase flow images, annotated into four regimes including Taylor bubble, bubbly, slug and churn. These regimes comprehensively capture diverse gas-liquid two-phase flow dynamics and regime transitions in a vertical flow setup. The data acquisition covers static and dynamic circulation scenarios with water and air injection and monitors key operating parameters, such as kick influx pressure $p_k$ (\si{psig}), circulation rate $Q_c$ (\si{gpm}), injection time $t_i$ (\si{s}), liquid flow rate $Q_l$ (\si{gpm}), liquid superficial velocity $u_l$ (\si{\foot\per\second}), gas flow rate $Q_g$ (\si{\cubicfoot\per\minute}), gas type $g_t$ (air or $\mathrm{CO_2}$) and floor number $f_n$.

\subsection*{Problem Settings}
\label{subsec: 4_2_Problem Settings}
In diagnosis tasks, a signal response sequence ${\mathbf{x}_{1:T}} = \{ {x_1},{x_2}, \ldots ,{x_T}\}  \in {{\mathbb{R}}^{T \times 1}}$ is collected from an industrial system with a corresponding set of physical-variable sequences ${\mathbf{V}_{1:{T_p}}} = \{ \mathbf{v}_{1:{T_p}}^1,\mathbf{v}_{1:{T_p}}^2, \ldots ,\mathbf{v}_{1:{T_p}}^C\}  \in {\mathbb{R}^{{T_p} \times C}}$, where $x_i$ denotes the value at the $i$-th time point, $\mathbf{v}_{1:{T_p}}^i$ represents the $i$-th physical variable, $C$ is the number of physical variables, $T$ and $T_p$ are the observation lengths of the signal response and the physical variables, respectively. It is worth noting that any variable capable of providing effective physical constraints or causal information for intensity diagnosis can be used as a physical variable and its temporal length is not required to be identical to the signal response length, i.e. $T \ne {T_p}$. To capture the dynamic properties of the signal response under non-stationary operating conditions and construct subsequent fundamental analysis units, the original signal response sequence $\mathbf{x}_{1:T}$ is split into multiple subsequences ${\mathbf{X}_{1:W_s}} = \{ \mathbf{x}_{1:W_s}^1,\mathbf{x}_{1:W_s}^2, \ldots ,\mathbf{x}_{1:W_s}^M\}  \in {\mathbb{R}^{W_s \times M}}$ via a sliding-window strategy, and then each subsequence $\mathbf{x}^i_{1:W_s}$ is converted into a corresponding time-frequency spectrogram $\mathbf{s}^i \in \mathbb{R}^{H \times W}$. The above process is described as follows:
\begin{equation}
\label{eq:data_augmentation}
\begin{aligned}
\mathbf{S}
&= \{ \mathbf{s}^1, \mathbf{s}^2, \ldots, \mathbf{s}^M \}, \\
&= \mathcal{T}\!\left( \mathcal{W}(\mathbf{x}_{1:T}) \right), \\
&= \{ \mathcal{T}(\mathbf{x}_{1:W_s}^i) \}_{i=1}^{M},
\end{aligned}
\end{equation}
where $\mathcal{W}(\cdot)$ denotes the sliding-window operator with a window size of $W_s$ and a stride of $S$, $M = \left\lfloor {\frac{{T - W_s}}{S}} \right\rfloor  + 1$ is the number of subsequences, $\mathcal{T}(\cdot)$ represents the continuous wavelet transform operator, $\mathbf{x}_{1:W_s}^i$ is the $i$-th subsequence, $\mathbf{s}^i$ is the time-frequency representation of $\mathbf{x}_{1:W_s}^i$, and $H$ and $W$ are the height and width of the time-frequency spectrogram $\mathbf{s}^i$.

\subsection*{Response Embedding}
\label{subsec: 4_3_Response Embedding}
Most existing Transformer-based diagnosis models typically embed discrete time points or consecutive signal segments as temporal tokens and apply self-attention to model temporal dependencies. To finely capture the evolution characteristics of signal responses within the time-frequency domain, PhyNet adopts time-frequency spectrogram-based patch-wise representations. Specifically, each time-frequency spectrogram $\mathbf{s}^i$ is partitioned into non-overlapping patches $\mathbf{Z}= \{ \mathbf{z}^1, \mathbf{z}^2, \ldots, \mathbf{z}^N \}$ along the time-frequency plane. Each patch $\mathbf{z}^i$ is subsequently mapped into a time-frequency response token $\mathbf{R}_{re,i} \in \mathbb{R}^{D \times 1}$ via linear projection. Given the distinct semantic roles of physical variables ${\mathbf{V}_{1:{T_p}}}$ and signal responses ${\mathbf{x}_{1:T}}$ in intensity diagnosis tasks, PhyNet employs distinct granularity embedding strategies to model them. To avoid the information misalignment caused by directly fusing physical-variable tokens ${\mathbf{P}_{va}} = \{ \mathbf{P}_{va,i}\} _{i = 1}^C \in \mathbb{R}^{C \times D}$ and time-frequency response tokens ${\mathbf{R}_{re}} = \{ \mathbf{R}_{re,i}\} _{i = 1}^N \in \mathbb{R}^{N \times D}$, PhyNet introduces a learnable global response token $\mathbf{G}_{re}\in \mathbb{R}^{D \times 1}$ for each signal response, which aggregates the macroscopic time-frequency representation of the signal and serves as a bridge for interaction between the signal response domain and the physical variable domain. This design facilitates the effective incorporation of causal and constraint information from physical variables into the modeling of time-frequency response patches. Overall, time-frequency response tokens ${\mathbf{R}_{re}}$ and the learnable global response token $\mathbf{G}_{re}$ are concatenated and fed into the Transformer encoder. The overall response embedding is formally expressed as follows:
\begin{equation}
\label{eq: response embedding}
\begin{aligned}
\{ {\mathbf{z}^1},{\mathbf{z}^2}, \ldots ,{\mathbf{z}^N}\}  &= \text{Patchify}({\mathbf{s}^i}),\\
{\mathbf{R}_{re}} &= \text{PatchEmbed}({\mathbf{z}^1},{\mathbf{z}^2}, \ldots ,{\mathbf{z}^N}),\\
{\mathbf{G}_{re}} &= \text{Learnable}(\mathbf{x}),\\
[{\mathbf{R}_{re}},{\mathbf{G}_{re}}]& \to \text{Encoder}( \cdot ),
\end{aligned}
\end{equation}
where $\text{Patchify}(\cdot)$ denotes the operation of dividing each time-frequency spectrogram $\mathbf{s}^i$ into non-overlapping patches of size $h \times w$, resulting in a total of $N = \left\lfloor {H/h} \right\rfloor  \times \left\lfloor {W/w} \right\rfloor$ patches. $\text{PatchEmbed}(\cdot)$ maps each $h \times w$ patch, combined with its positional embedding, into a $D$-dimensional vector via a learnable linear projection. $[\cdot,\cdot]$ represents the concatenation operation, and $\text{Encoder}( \cdot )$ denotes the Transformer encoder.

\subsection*{Physical Variables Embedding}
\label{subsec: 4_4_Physical Variables Embedding}
The primary role of physical variables is to provide effective physical constraints and causal information for intensity diagnosis. In real-world industrial systems, different physical variables exhibit different sampling frequencies and observation lengths. To naturally capture the interactions among physical variables while avoiding the computational overhead and unnecessary noise introduced by overly fine-grained representations, PhyNet embeds each physical variable sequence $\mathbf{v}_{1:{T_p}}^i$ as a sequence-wise variate token $ \mathbf{P}_{va,i} \in \mathbb{R}^{D \times 1}$. This design can adapt to various observation lengths and irregularities, providing a robust global representation of physical information for subsequent interaction with the time-frequency response tokens ${\mathbf{R}_{re}}$. This process is formally represented as follows:
\begin{equation}
\label{eq: physical variables embedding}
\begin{aligned}
{\mathbf{P}_{va}} = \{ \mathbf{P}_{va,i}\} _{i = 1}^C &= \text{VariateEmbed}(\mathbf{v}_{1:{T_p}}^i),\\
{\mathbf{P}_{va}} & \to \text{Encoder}( \cdot ),
\end{aligned}
\end{equation}
where $\text{VariateEmbed}(\cdot)$ maps each physical variable sequence $\mathbf{v}_{1:{T_p}}^i$ into a $D$-dimensional embedding vector via a trainable linear projection, $C$ is the number of physical variables, $T_p$ is the observation length of the physical variables, and $\text{Encoder}( \cdot )$ denotes the Transformer encoder.

\subsection*{Response Self-Attention}
\label{subsec: 4_5_Response Self-Attention}
In the intensity diagnosis task, to effectively model the time-frequency dependencies of signal response sequences and their global interactions with physical variable sequences, PhyNet introduces a self-attention mechanism over the time-frequency response tokens ${\mathbf{R}_{re}}$. Concretely, self-attention among local time-frequency response tokens ${\mathbf{R}_{re}}$ is employed to capture fine-grained time-frequency relationships within the signal response sequence, i.e. ${\rm{Local}} \to {\rm{Local}}$, while a learnable global response token ${\mathbf{G}_{re}}$ serves as a bridge between the signal response sequence ${\mathbf{x}_{1:T}}$ and the physical variable sequence ${\mathbf{V}_{1:{T_p}}}$. In cross-domain attention, the global response token ${\mathbf{G}_{re}}$ plays an asymmetric role. On the one hand, it aggregates local time-frequency patch information across the entire signal response sequence to form a macroscopic representation of the signal response sequence, i.e. ${\rm{Local}} \to {\rm{Global}}$. On the other hand, it provides each local time-frequency token with global constraints and correlational information derived from the physical variables, i.e. ${\rm{Global}} \to {\rm{Local}}$. This mechanism not only enables a comprehensive characterization of the dynamic evolution of signal responses, but also enhances the interaction with irregular physical-variable sequences. The attention mechanism can be expressed as follows:
\begin{equation}
\label{eq: response self-attention}
\begin{aligned}
\text{Local} \rightarrow \text{Local}{\rm{:}}\; \tilde{\mathbf{R}}_{re}^{l,1} &= \mathrm{LayerNorm}(\mathbf{R}_{re}^{l}+ \mathrm{Self\text{-}Attention}(\mathbf{R}_{re}^{l})),\\
\text{Global} \rightarrow \text{Local}{\rm{:}}\;\tilde{\mathbf{R}}_{re}^{l,2} &= \mathrm{LayerNorm}(\mathbf{R}_{re}^{l}+ \mathrm{Cross\text{-}Attention}(\mathbf{R}_{re}^{l}, \mathbf{G}_{re}^{l})),\\
\text{Local} \rightarrow \text{Global}{\rm{:}}\; \tilde{\mathbf{G}}_{re}^{l} &= \mathrm{LayerNorm}(\mathbf{G}_{re}^{l}+ \mathrm{Cross\text{-}Attention}(\mathbf{G}_{re}^{l}, \mathbf{R}_{re}^{l})).
\end{aligned}
\end{equation}
The entire process can be further expressed as a single response self-attention computation:
\begin{equation}
\label{eq: simple response self-attention}
\tilde{\mathbf{R}}_{re}^{l}, \tilde{\mathbf{G}}_{re}^{l}  = \mathrm{LayerNorm}([\mathbf{R}_{re}^{l},\mathbf{G}_{re}^{l}] +\mathrm{Self\text{-}Attention}([\mathbf{R}_{re}^{l},\mathbf{G}_{re}^{l}])),
\end{equation}
where $l \in \{ 0,1, \ldots ,L - 1\}$ denotes the $l$-th encoder layer of PhyNet, with $\mathbf{R}_{re}^{0}=\mathbf{R}_{re}$, $\mathbf{G}_{re}^{0}=\mathbf{G}_{re}$. $[\cdot,\cdot]$ represents the concatenation of patch-wise time-frequency response tokens ${\mathbf{R}_{re}}$ and the global response token ${\mathbf{G}_{re}}$ along the sequence dimension. By applying a self-attention mechanism to the concatenated response tokens $[\mathbf{R}_{re}^{l},\mathbf{G}_{re}^{l}]$, PhyNet can simultaneously model the local dependencies among different time-frequency patches and the global relationships between each patch and the entire signal response sequence, achieving a holistic characterization of the dynamic evolution of signal responses.

\subsection*{Physical-to-Response Cross-Attention}
\label{subsec: 4_6_Physical-to-Response Cross-Attention}
To fully integrate the global constraints and causal information provided by the physical-variable sequences ${\mathbf{V}_{1:{T_p}}}$, PhyNet introduces a cross-attention mechanism between the time-frequency response tokens ${\mathbf{R}_{re}}$ and the physical-variable tokens ${\mathbf{P}_{va}}$. Specifically, the cross-attention layer takes the time-frequency response tokens ${\mathbf{R}_{re}}$ as queries and the physical-variable tokens ${\mathbf{P}_{va}}$ as keys and values, establishing connections between the signal response and the physical constraints. Since the physical variables are embedded as sequence-wise variate tokens, the learnable global response token is used to aggregate global information from the physical variables. The above process can be formalized as:
\begin{equation}
\label{eq: physical-to-response cross-attention}
\tilde{\mathbf{G}}_{re}^{l}= \mathrm{LayerNorm}(\tilde{\mathbf{G}}_{re}^{l}+ \mathrm{Cross\text{-}Attention}(\tilde{\mathbf{G}}_{re}^{l}, {\mathbf{P}_{va}})),
\end{equation}
Subsequently, all local time-frequency response tokens ${\mathbf{R}}_{re}$ and the learnable global response token $\mathbf{G}_{re}$ are updated through a feed-forward network, which can be formally expressed as follows:
\begin{equation}
\label{eq: feed-forward}
\begin{aligned}
\mathbf{R}_{re}^{l+1}=\mathrm{Feed\text{-}Forward}(\tilde{\mathbf{R}}_{re}^{l}),\\
\mathbf{G}_{re}^{l+1}=\mathrm{Feed\text{-}Forward}(\tilde{\mathbf{G}}_{re}^{l}),\\
\end{aligned}
\end{equation}
where $l \in \{1, \ldots ,L\}$. This design allows the learnable global response token to simultaneously integrate local patch-wise information and global physical constraints, enabling effective interactions between signal responses and physical variables.

\subsection*{Optimization Objective}
\label{subsec: 4_7_Optimization Objective}
In intensity diagnosis tasks, the physical variables ${\mathbf{V}_{1:{T_p}}}$ serve as auxiliary information and are not required to be predicted. Therefore, the diagnostic output is generated by performing a linear projection onto the final response embeddings produced by PhyNet. Specifically, the patch-wise time-frequency response tokens ${\mathbf{R}_{re}}$ and the global response token ${\mathbf{G}_{re}}$ from the last encoder layer of PhyNet are aggregated and then passed through a LayerNorm operation followed by a linear projection to obtain the predicted probability distribution over intensity classes. To measure the discrepancy between the predicted probabilities ${{\hat y}_i}$ and the ground-truth intensity labels ${y_i}$, PhyNet adopts the cross-entropy loss function for supervised learning, which is defined as:
\begin{equation}
\mathcal{L} = -\sum\nolimits_{i = 1}^I {{y_i}\log } ({{\hat y}_i}),\;\; \text{where}\;\;{{\hat y}_i} = \text{Projection}(\text{LayerNorm}[\mathbf{R}_{re}^L,\mathbf{G}_{re}^L]),
\end{equation}
where $I$ denotes the number of intensity classes, $L$ represents the index of the last encoder layer and $[\cdot,\cdot]$ denotes the concatenation operation.

\subsection*{Baselines}
\label{subsec: 4_8_Baselines}
To validate the effectiveness and advantages of the data-physics hybrid PhyNet model, we categorize baselines into three groups based on their utilization of data and physical variables. This categorization enables a systematic evaluation of the benefits of integrating physical information with signal-based representations for intensity diagnosis.

\noindent\textbf{Data-driven methods.} Data-driven methods commonly adopt end-to-end deep learning approaches, which are organized into three types: CNN-based methods, Transformer-based methods and CNN-Transformer hybrid methods. In detail, CNN-based methods excel at capturing local patterns from signals and include models such as ResNet \cite{he2016deep}, DenseNet \cite{huang2017densely} and ConvNeXt \cite{liu2022convnet}. Transformer-based methods are effective at modelling long-term dependencies from signals, including models such as ViT \cite{dosovitskiy2020vit}, Swin \cite{liu2021swin}, PerViT \cite{min2022peripheral} and CrossViT \cite{chen2021crossvit}. CNN-Transformer hybrid methods combine local feature modeling of CNNs with the global dependency modelling of Transformers, including models such as UniFormer \cite{li2023uniformer}, ConFormer \cite{49414}, CFFormer \cite{li2026cfformer} and LeViT \cite{graham2021levit}. For all data-driven models, original sensing measurements are directly used as inputs, without explicit incorporation of physical variables, rendering these models purely data-driven baselines.

\noindent\textbf{Physics-driven methods.} Physics-driven methods essentially use traditional machine-learning algorithms with physical variables as input features for classification. Specifically, support vector machines (SVM) \cite{cortes1995support} classify samples by finding the optimal hyperplane that separates different classes. Decision Trees (DT) \cite{quinlan1986induction} recursively split the feature space based on input features to generate decision rules. Extreme Gradient Boosting (XGBoost) \cite{chen2016xgboost} builds an ensemble of gradient-boosted trees to improve predictive performance. The K-nearest neighbors (KNN) algorithm \cite{fukunaga2013introduction} assigns class labels based on the majority label among the nearest neighbors in the feature space.

\noindent\textbf{Data-Physics Hybrid methods.} Data-physics hybrid methods are a class of multimodal learning methods that jointly process multiple heterogeneous data sources to capture complementary information for improved representation learning. Concretely, Vision-and-Language Transformer (ViLT) \cite{pmlrv139kim21k} uses a Transformer to encode and align visual and tabular inputs in a unified latent space. Perceiver \cite{jaegle2021perceiver} processes high-dimensional inputs by iteratively mapping them into a fixed-size latent array through cross-attention, enabling flexible multimodal representation. PerceiverIO (PIO) \cite{jaegle2021perceiverio} extends the Perceiver architecture to process arbitrary multimodal inputs through iterative cross-attention between latent arrays and input modalities. Multimodal Transformer (MulT) \cite{tsai2019multimodal} employs cross-modal attention layers to explicitly model interactions between different modalities and fuse information across them. For all data-physics hybrid methods, both original sensing measurements and physical variables are used as inputs, enabling the models to jointly learn from multiple modalities.

\subsection*{Experimental Settings and Reproducibility}
\label{subsec:4_9_experimental_settings}
All experiments are conducted on a Linux workstation equipped with two NVIDIA GeForce RTX 5090 GPUs, using Python 3.10.19, PyTorch 2.8.0, CUDA 12.9 and cuDNN 9.10.2. All models are trained from scratch with randomly initialized parameters and without external pre-trained weights. DenseNet models use an input resolution of $256 \times 256$, whereas all other models use an input resolution of $224 \times 224$. All deep learning models are optimized using the cross-entropy loss and the Adam optimizer with $\beta_1 = 0.9$, $\beta_2 = 0.999$, $\varepsilon = 10^{-8}$ and zero weight decay. The initial learning rate is set to $1 \times 10^{-4}$, with a batch size of 16 and a cosine learning-rate decay schedule. To improve training stability, the maximum gradient norm is clipped to 4.0. All models are trained for 30 epochs without early stopping. The dropout settings of the baseline models follow their respective implementations. The default PhyNet configuration uses three input channels, a patch size of $16 \times 16$, a 512-dimensional embedding, 8 attention heads, 3 encoder layers, a feed-forward dimension of 2048, a dropout rate of 0.1 and the GELU activation function. All models are trained and evaluated using predefined training and test splits. Further implementation details are provided in the Supplementary Methods.

\subsection*{Metrics}
\label{subsec: 4_9_Metrics}
For intensity diagnosis tasks, model performance is evaluated using a comprehensive set of metrics. We report accuracy (Acc.), precision (Pre.), recall (Rec.) and F1-score (F1) to assess the model's ability to correctly identify all classes, particularly in imbalanced scenarios \cite{rainio2024evaluation,deltadahl2025deep}. The area under the receiver operating characteristic curve (AUC-ROC) provides a threshold-independent evaluation of discriminative performance \cite{korot2021code}. The confusion matrix is also examined to visualize class-wise prediction patterns and common misclassifications \cite{carnevali2024deep}. All metrics are calculated on held-out test sets to ensure a robust and comprehensive evaluation of classification performance. Detailed definitions and calculation procedures for these metrics are provided in the Supplementary Information.
\section*{Data availability}
\label{sec: 5_Data availability}
The cavitation dataset is available via Dropbox at \href{https://www.dropbox.com/scl/fo/92pohahsa4uxezd9qqd28/ANSRMT-zca-5_zYfDEPKpE0?rlkey=8eiyb3sbnpz1jel6gzuw0ameq&st=lmbu3en4&dl=0}{https://www.dropbox.com/data/cavitation}. The other dataset used in this study is publicly available and can be accessed as described in \cite{manikonda2025vertical}. Source data are provided with this paper.

\section*{Code availability}
\label{sec: 6_Code availability}
The code for PhyNet is publicly available via GitHub at \href{https://github.com/CavitationDetection/PhyNet}{https://github.com/CavitationDetection/PhyNet}.
\bibliography{sn-bibliography}
\section*{Acknowledgements}
\label{sec: 7_Acknowledgements}
This research is supported by the China Postdoctoral Science Foundation under Grant No.\ 2025M781519 (Y.S.), the National Natural Science Foundation of China under Grant No.\ 92570117, the Ministry of Science and Technology of China under Grant No.\ 2024YFA1611004, the Shenzhen Peacock Fund under Grant Nos. 2023TC0007 and 2023TC0179, the CUHK-Shenzhen University Development Fund under Grant Nos. UDF01003041 and UDF03003041 (Y.S., K.Z.), the AI grant at FIAS through SAMSON AG (K.Z.), the BMBF funded KISS consortium (05D23RI1) in the ErUM-Data action plan (K.Z.), the SAMSON AG (D.V., A.W.), the Walter Greiner Gesellschaft zur F\"orderung der physikalischen Grundlagenforschung e.V. through the Judah M. Eisenberg Laureatus Chair at Goethe Universit\"at Frankfurt am Main (H.S.) and the NVIDIA GPU grant through NVIDIA Corporation (K.Z.).
\section*{Author contributions}
\label{sec: 8_Author contributions}
Y.S. and K.Z. conceived and designed the research. Y.S. conducted the research and drafted the original manuscript. J.T., D.Z., H.S., D.V. and A.W. contributed to the analysis, discussion and validation. D.V. and A.W. provided the cavitation data. Y.S., H.S., D.V., A.W. and K.Z. acquired funding for the project. K.Z. supervised the study. All authors read and approved the final manuscript.

\section*{Competing interests}
\label{sec: 9_Competing interests}
D.V. and A.W. are employees of SAMSON AG, which provided the cavitation datasets used in this study. K.Z. received research support from SAMSON AG. The remaining authors declare no competing interests.

\fi

\end{document}